\documentclass[final, 3p, times]{elsarticle}

\usepackage{lineno,hyperref}
\modulolinenumbers[5]

\journal{Information Sciences}

\usepackage{amsmath}
\usepackage{array}
\usepackage{graphicx}
\usepackage{subfloat}
\usepackage{multirow}
\usepackage{subfig}
\usepackage{makecell}
\usepackage{times}
\usepackage{latexsym}
\usepackage{booktabs} 
\usepackage{multirow}
\usepackage{url}
\usepackage{amsmath}
\usepackage{bm}
\usepackage{amsfonts}
\usepackage{graphicx}
\usepackage{diagbox}
\usepackage{xspace}
\usepackage{courier}
\usepackage{mathrsfs}
\usepackage{float}
\usepackage{framed} 
\usepackage{color}
\usepackage{amsfonts}
\usepackage{amssymb}
\usepackage{diagbox}
\usepackage{subfig}
\usepackage{array}
\hypersetup{hidelinks}
\usepackage[normalem]{ulem}
\usepackage{bm}
\usepackage{makecell}
\usepackage{multirow}
\usepackage{url}
\usepackage{graphicx}
\usepackage{appendix}
\usepackage{float}
\usepackage{color}
\usepackage{subfig}
\usepackage{array}
\usepackage{color}
\useunder{\uline}{\ul}{}
\usepackage{amsmath}
\usepackage{array}
\usepackage{graphicx}
\usepackage{subfloat}
\usepackage{multirow}
\usepackage{subfig}
\hypersetup{hidelinks}
\biboptions{sort&compress} 
 
\usepackage[labelfont=bf]{caption}  
 
\makeatletter 
\def\@makecaption#1#2{%
	\vskip\abovecaptionskip 
	\sbox\@tempboxa{#1 #2}%
	{\bfseries #1} #2\par 
	\vskip\belowcaptionskip} 
\makeatother 

\usepackage{color}
\definecolor{REVISE}{RGB}{0, 0, 255}

\begin{document}

\begin{frontmatter}

\title{Contextual Dictionary Lookup for Knowledge Graph Completion}

\author[mymainaddress]{Jining~Wang}
\author[hismainaddress]{Delai~Qiu}
\author[mymainaddress]{YouMing~Liu}
\author[hismainaddress]{Yining~Wang}
\author[mymainaddress]{Chuan~Chen\corref{mycorrespondingauthor}}
\cortext[mycorrespondingauthor]{Corresponding author}
\ead{chenchuan@mail.sysu.edu.cn}

\author[mymainaddress]{Zibin~Zheng}

\author[mymainaddress]{Yuren~Zhou}

\address[mymainaddress]{School of Computer Science and Engineering, Sun Yat-sen University, Guangzhou, 510275, China}
\address[hismainaddress]{Unisound Information Technology Co., Ltd, Beijing, 100089, China}

\begin{abstract}
	Knowledge graph completion (KGC) aims to solve the incompleteness of knowledge graphs (KGs) by predicting missing links from known triples, numbers of knowledge graph embedding (KGE) models have been proposed to perform KGC by learning embeddings. Nevertheless, most existing embedding models map each relation into a unique vector, overlooking the specific fine-grained semantics of them under different entities. Additionally, the few available fine-grained semantic models rely on clustering algorithms, resulting in limited performance and applicability due to the cumbersome two-stage training process. In this paper, we present a novel method utilizing contextual dictionary lookup, enabling conventional embedding models to learn fine-grained semantics of relations in an end-to-end manner. More specifically, we represent each relation using a dictionary that contains multiple latent semantics. The composition of a given entity and the dictionary's central semantics serves as the context for generating a lookup, thus determining the fine-grained semantics of the relation adaptively. The proposed loss function optimizes both the central and fine-grained semantics simultaneously to ensure their semantic consistency. Besides, we introduce two metrics to assess the validity and accuracy of the dictionary lookup operation. We extend several KGE models with the method, resulting in substantial performance improvements on widely-used benchmark datasets. 
\end{abstract}

\begin{keyword}
	Knowledge Graph Completion \sep Fine-grained Semantics \sep Contextual Dictionary Lookup
\end{keyword}

\end{frontmatter}

\section{Introduction}

\begin{figure*}[!t]
\centering
\subfloat[]{\includegraphics[width=0.46\hsize]{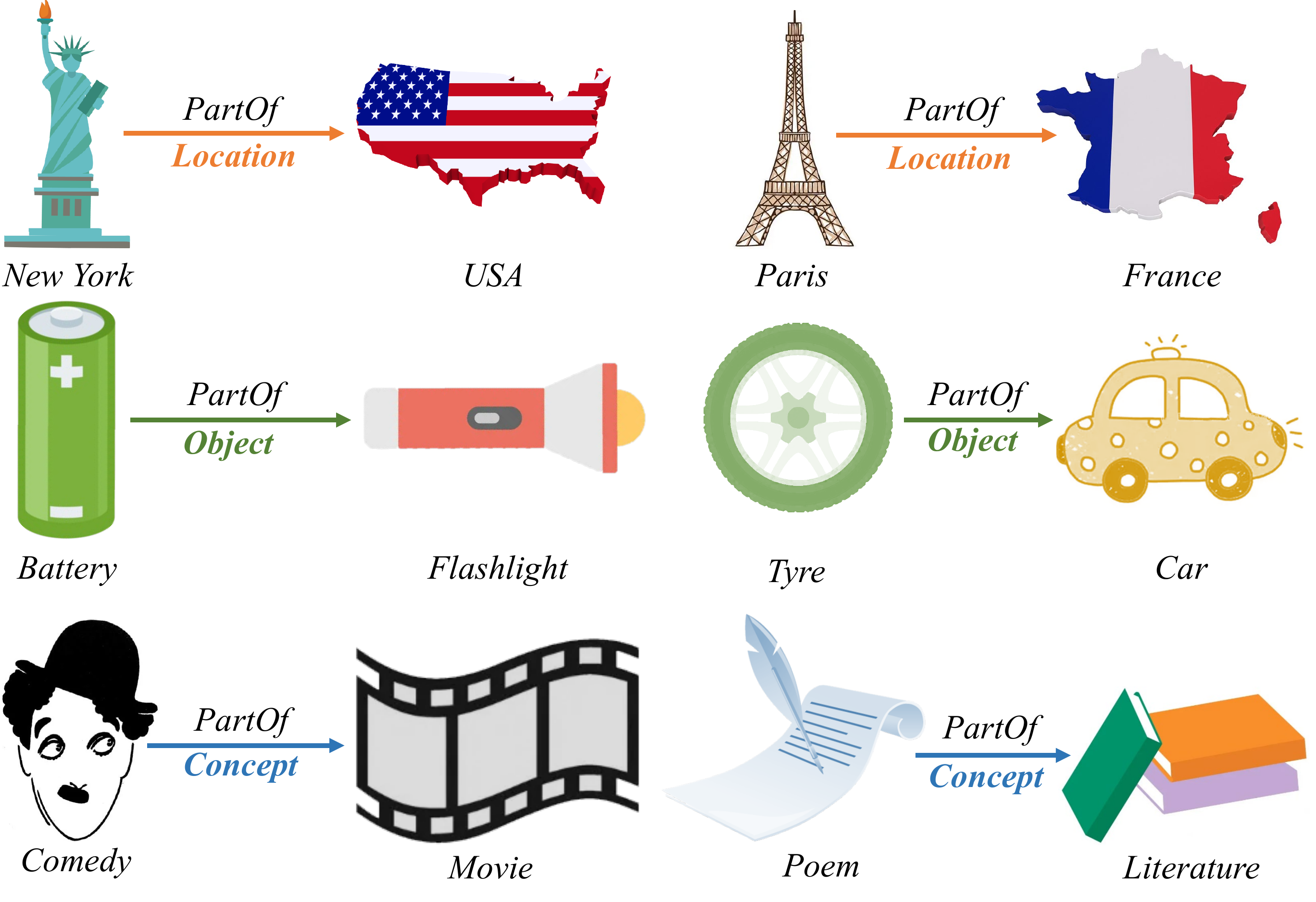}}
\subfloat[]{\includegraphics[width=0.54\hsize]{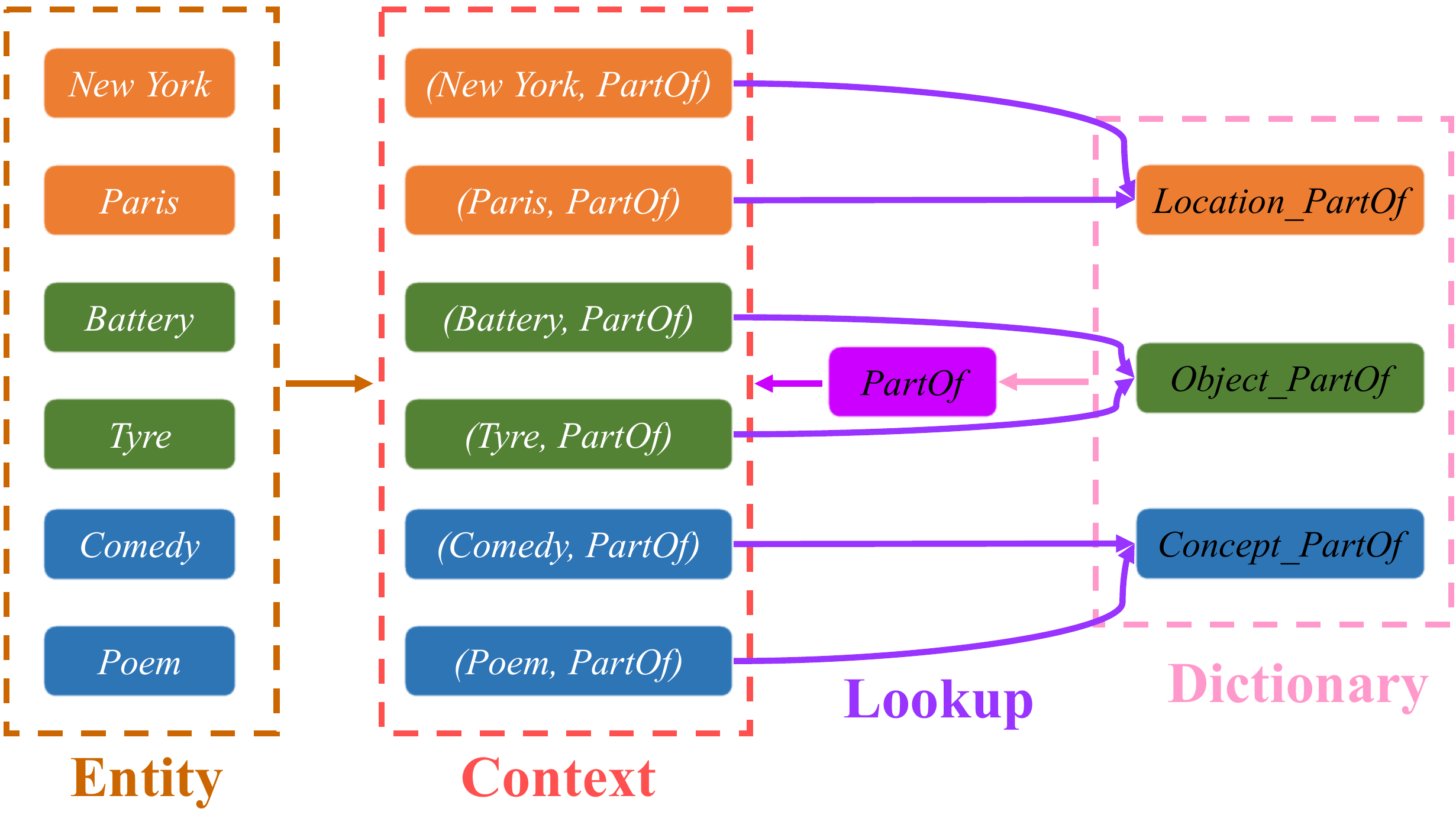}}
\caption{Overview of how CoDLR works to obtain the fine-grained semantics of relations. (a) The relation \textit{PartOf} has diverse fine-grained semantics under different entities, which are marked below the arrows of corresponding colors. (b) The calculation diagram of CoDLR, the target relation \textit{PartOf} is represented with a dictionary containing amount of semantics, the context combines both different entities and central semantics of the dictionary, which is used to generate a lookup on the dictionary to determine the choice of fine-grained semantics.}
\label{example}

\end{figure*}

Knowledge graphs (KGs) are structured representations of factual triplets in the form of multi-relational graphs. Each triplet $(h, r, t)$ consists of a head entity, a tail entity, and the relation that connects them. KGs play a crucial role in organizing human knowledge and support applications in various downstream tasks of natural language processing (NLP) such as question answering \cite{qatask}, dialog systems \cite{dstask}, information retrieval \cite{irtask}, and recommendation systems \cite{rstask}. Despite containing millions of triples, KGs are inherently incomplete and cannot capture all possible facts \cite{incompleteness}. This limitation has led to the development of knowledge graph completion (KGC) algorithms.

KGC, also known as link prediction, aims to automatically predict missing triples in KGs. Concretely, the task can be regarded as question answering, i.e., given an incomplete triple like $(h, r, ?)$ or $(?, r, t)$, the objective is to infer suitable tail (or head) entities from a set of candidate options. There are a lot of knowledge graph
embedding (KGE) models proposed for KGC. These models share a common approach of projecting entities and relations from KGs into a lower-dimensional continuous space. Specifically, additive models \cite{transe, transh, transr, transd} take relations as transfer vectors to convert head entities to tail entities through addition operation, while multiplicative models \cite{rescal, distmult, complex, quate} measure the plausibility of candidates with multiplication-related semantic similarity-based score function. To capture deeper nonlinear interactions among entities and relations, recent studies have focused on constructing innovative model structures by leveraging various neural networks \cite{nam, conve, interacte, rgcn}. 

While KGE models have achieved remarkable success in KGC, they often fall short in capturing the fine-grained semantics of relations. We define the semantics of a relation as its specific meaning in a triple, and most methods assume that the semantics of a relation is singular and static. However, a relation may have multiple specific fine-grained semantics across different triples in reality. For example, consider the relation \textit{PartOf} shown in Fig.~\ref{example}(a). When it connects with the entity \textit{New York}, the fine-grained semantics related to \textit{location} is expressed, and semantics related to \textit{object} or \textit{concept} can be observed under other entities. Although a few models have been proposed to address this issue and utilize fine-grained semantics for improved expression ability, they often rely on clustering-based two-stage training methods \cite{transr, transd}, which have notable limitations. 

CTransR \cite{transr} serves as an example of such models to explain the training process. In the first stage, a well-trained TransE model \cite{transe} is required, which is expected to generate embeddings $(\bm{h}, \bm{r}, \bm{t})$ for triple $(h, r, t)$ with the goal that $\bm{h}+\bm{r} \approx \bm{t}$. Subsequently, the vector offsets $\bm{h}-\bm{t}$ are clustered into several groups with different semantics. In the second stage, a separate relation vector $\bm{r_c}$ is learned for each group and used to replace $\bm{r}$ to train a final model that considers the fine-grained semantics. Apparently, the two-stage process increases the difficulty and workload of training. Apart from this, to obtain relation with fine-grained semantics from head entity and tail entity in the opposite direction, only models using reversible additive operation like TransE can be considered in the first stage. However, due to the difference of embedding distribution learned by different models \cite{dog}, initializing complex models with embeddings from simple models directly in the second stage will cause the decline of expression ability. Besides, the semantics of an ambiguous relation can be viewed as a combination of a few fine-grained semantics. Unfortunately, the aforementioned models are incapable of properly handling such cases since they classify the semantics of a given relation into a certain cluster.

To overcome above disadvantages, we propose a new method called \textbf{CoDLR} (\textbf{Co}ntextual \textbf{D}ictionary \textbf{L}ookup for \textbf{R}elations). Compared with the previous methods, CoDLR offers significant improvements in two key aspects: \textbf{(1) CoDLR preserves the expression ability of the original models.} Instead of extracting the fine-grained semantics of relations from a well-trained model, CoDLR obtains them by learning the dictionary from scratch. This approach ensures that the distribution of embeddings remains unchanged and unaffected by other models introduced in the first stage of a cluster-based scheme. \textbf{(2) CoDLR provides better representation of the ambiguous semantics of relations.} Unlike previous models that classify fine-grained semantics into specific clusters, CoDLR combines the semantics within the dictionary using a lookup vector. This allows for more flexible modeling of ambiguous relations.

We provide an illustration for CoDLR in Fig.~\ref{example}(b), given one triple or query, the relation is represented by a dictionary containing fine-grained semantics in several aspects. To determine the most suitable semantics, we consider the composition of the relation-connected entity and the central semantics of the dictionary as the context associated with the current fine-grained semantics. The context is used to generate a lookup vector consisting of probabilities, which is then multiplied with the dictionary. For example, if the question to be answered is \textit{(New York, PartOf, ?)}, we compose the representation of \textit{New York} and the central semantics of the dictionary about \textit{PartOf} to obtain the context. Note that the composition operation can be implemented in various ways to ensure the flexibility and expressive power of the model. Then, the lookup vector generated from the context vector can help to select \textit{Location\_PartOf} in the dictionary as appropriate fine-grained semantics of \textit{PartOf} under the entity \textit{New York}. To maintain the consistency between the central semantics and fine-grained semantics of the dictionary, we optimize two loss functions containing them respectively in parallel during the training process. Besides, we measure the quality of dictionary lookup in terms of validity and accuracy with metrics called \textbf{SOL} (\textbf{S}parseness \textbf{O}f \textbf{L}ookup) and \textbf{DAE} (\textbf{D}ictionary \textbf{A}lignment \textbf{E}ntity), where SOL checks if a lookup vector is sparse enough to focus on the most appropriate semantics, DAE inspects if the diversity of dictionary aligns to that of corresponding entities.

In summary, our contributions in this paper are the following:

\begin{itemize}
\item We propose a contextual dictionary lookup based method called CoDLR to enable the end-to-end fine-grained semantics discovery for various kinds of KGE models, which is the first study in this field to the best of our knowledge.

\item We propose two metrics called SOL and DAE to measure the validity and accuracy of dictionary lookup respectively, which are applied to ensure the reliability of CoDLR.

\item We extend different existing KGE models including TransE, DistMult, and ConvE with CoDLR, the enhanced models outperform the original models as well as other competitors by a significant margin on popular benchmark datasets.
\end{itemize}

\section{Related work}
\label{rw}

In this section, we will provide an overview of popular KGE models, which include additive models and their variants incorporating relations with fine-grained semantics. We will also discuss multiplicative models and recent studies that have leveraged various neural networks for enhancement.

\subsection{Additive models}

Additive models embed entities and relations into vector space, then optimize the distance-based score function with the goal of making the head entity added by relation close to the tail entity. TransE \cite{transe} is the first study proposed in this field, where embeddings of entities and relations are translated in the same space. To model complex relation patterns including one-to-many, many-to-one, and many-to-many, subsequent works assume that entities and relations should be distributed in different spaces. For example, TransH \cite{transh} obtains various entity representation under different relations by projecting the entity embeddings into the relation-specific hyperplane, while TransR \cite{transr} replace the hyperplane with the semantic space possessing stronger expression, and TransD \cite{transd} introduces entity-specific vectors for projection operation so that the attributes of head and tail entities can be considered in addition to that of relations.

The calculation direction of the above studies is from relation to entity, i.e., extracting corresponding attributes of entities under given relations. In the opposite direction, relations also have fine-grained semantics under specific entities. Benefiting from the reversibility of addition, CTransR \cite{transr} and HRS \cite{hrs} cluster the vector offsets between head entities and tail entities obtained from well-trained TransE into a plurality of groups at first, the center of each group corresponds to a fine-grained semantics of original relation, which can be applied to other models in the second training stage. The difference between the two models lies in that the former directly replaces the original relation with the fine-grained semantic one, while the latter adds them together to obtain the new representation.

\subsection{Multiplicative models}

Multiplicative models use the semantic similarity-based score function to measure the plausibility of triplets, enabling them to maintain good expression ability when the scale of KGs grows. Specifically, RESCAL \cite{rescal} and DistMult \cite{distmult} introduce bilinear transformation operations in point-wise space, while Complex \cite{complex} and RotatE \cite{rotate} represent embeddings in complex-valued space so that relations' complicated patterns like composition and antisymmetry can be captured.

\subsection{Neural network based models}

To capture complex interactions among triples, much attention has been paid to various types of neural networks. Considering that the simplicity and expressiveness of multi-layer perception (MLP) are suitable for enhancing score functions, NTN \cite{ntn}, SME \cite{sme} and NAM \cite{nam} explore nonlinear modeling with fully-connected layers and activation function. Convolutional neural networks have also been employed to improve expression ability by learning deeper features with fewer shared parameters. For instance, ConvE \cite{conve} and ConvKB \cite{convkb} apply filters over the matrix reshaped from embeddings of entities and relations, while ConvR \cite{convr} and InteractE \cite{interacte} propose innovative modules to maximize the number of interactions between entity and relation features. To utilize the graph structure information, graph neural networks are introduced by R-GCN \cite{rgcn}, SACN \cite{sacn} and VR-GCN \cite{vrgcn} for graph context modeling under the encoder-decoder framework.

\section{PRELIMINARIES}
In this section, several important concepts and notations related to our work will be introduced, the formal definition of KGs and KGC will be provided.

\subsection{Knowledge Graphs (KGs)}
We denote a knowledge graph as $\mathcal{G}=\{(h, r, t) \!\! \mid \!\! h, t \in \mathcal{E}, r \in \mathcal{R}\}$, which is a set of triples indeed, where $h, t$ correspond to head and tail entity, $r$ is the relation connecting both $h$ and $t$, $\mathcal{E, R}$ represent the set of entities and relations respectively.

\subsection{Knowledge Graph Completion (KGC)}

Given one question $(h, r, ?)$ or $(?, r, t)$, the KGC or link prediction aims to infer missing $t/h$ on $\mathcal{G}$ based on known triples. Let $\bm{h}, \bm{r}, \bm{t} \in R^d$ be the vector representation of the triple $(h, r, t)$ with dimension $d$ learned by KGE models, the score function $\psi: {R}^{d} \times {R}^{d} \times {R}^{d} \rightarrow {R}$ takes $\bm{h}, \bm{r}, \bm{t}$ as input and outputs a score measuring the plausibility of triplet, then candidate with the greatest possibility will be chosen as the final result. Note that the score function $\psi$ can be implemented in several ways, but the common goal is to score the valid triples higher than the invalid ones, three representative KGE models we extend will be reviewed as follows.

\subsubsection{TransE \cite{transe}} TransE defines distance-based score function in the pointwise Euclidean space, where $||$ means $L_2$ distance.

\begin{equation}
\psi(\bm{h},\bm{r},\bm{t}) = ||\bm{h}+\bm{r}-\bm{t}||
\label{etranse}
\end{equation}

\subsubsection{DistMult \cite{distmult}} DistMult defines multiplication-based score function as follows, note that $\bm{R}$ is a diagonal matrix generated from $\bm{r}$, and $\circ$ denotes Hadamard product.

\begin{equation}
\psi(\bm{h},\bm{r},\bm{t}) = \bm{h}^T\bm{R}\bm{t} = (\bm{h}\circ\bm{r})^T\bm{t}
\label{edist}
\end{equation}

\subsubsection{ConvE \cite{conve}} ConvE applies convolutional neural networks to define score function as follows, where $\sigma$ corresponds to the activation function, $\omega$ means convolution kernels, $*$ denotes 2D convolution, and $W$ is the learnable weight matrix.
\begin{equation}
\psi(\bm{h},\bm{r},\bm{t}) = \sigma(\operatorname{vec}(\sigma([\bm{h}, \bm{r}] * \omega)) W)^{\top} \bm{t}
\end{equation}

\section{Proposed approach}

\begin{figure*}[!t]
	\centering
	\includegraphics[width=1.0\hsize]{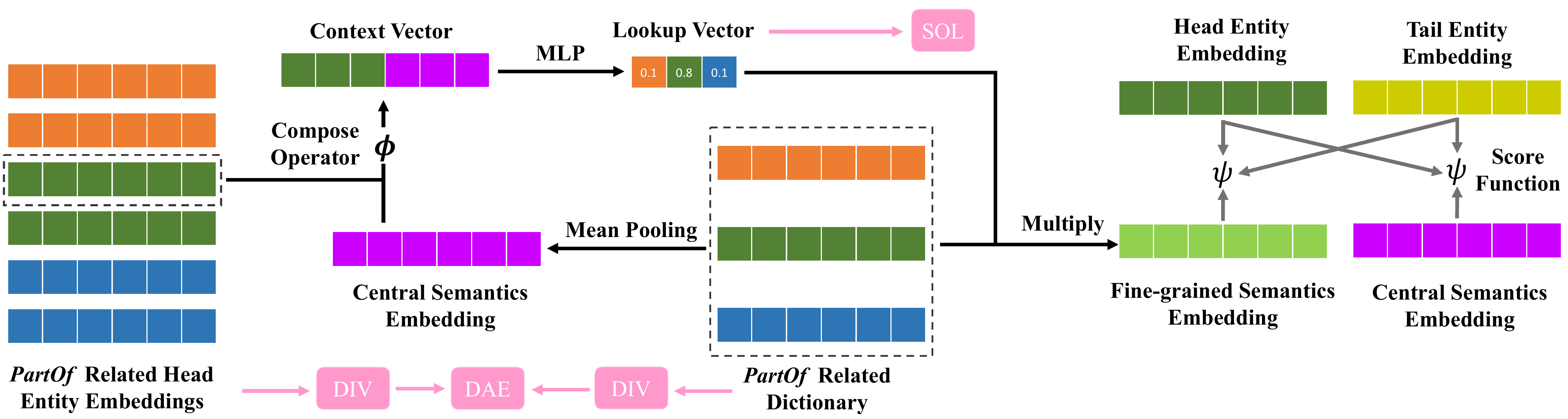}
	\caption{A detailed illustration for the proposed CoDLR as well as two metrics. Take \textit{PartOf} related questions as an example, the CoDLR begins by obtaining the central semantics embedding via mean pooling on the dictionary, which is composed with relation-connected entity embedding to get context vector. Then, the context vector generates a lookup vector to determine the fine-grained semantic embedding. In the end, both central and fine-grained semantics embeddings are utilized in score functions to evaluate the likelihood that the provided tail entity is the correct answer. Additionally, CoDLR employs two metrics for evaluation: SOL checks the sparsity of lookup vector, while DAE inspects whether corresponding entity embeddings are aligned to the dictionary.}
	\label{model}
\end{figure*}

This section presents CoDLR, an end-to-end framework designed to obtain the fine-grained semantics of relations through context dictionary lookup. Fig.~\ref{model} provides an illustration of the core idea behind CoDLR, and we will detail the framework along with two associated metrics in the rest.

\subsection{Dictionary Initialization}

Given triple \textit{(h, r, t)}, suppose the embedding dimension is $d$, the size of dictionary is $n$, CoDLR initializes \textit{h, t} with vectors $\bm{h}, \bm{t} \in R^d$, and maps \textit{r} to a trainable matrix $\bm{\mathcal{D}_r} \in R^{n \times d}$, which is a dictionary containing $n$ fine-grained semantics. Supposing the semantic center always coincides with the geometric center of $\bm{\mathcal{D}_r}$, we define the central semantics $\bm{\mathcal{D}^c_r} \in R^d$ as the mean vector of $\bm{\mathcal{D}_r}$, where $\bm{\mathcal{D}^i_r}\in R^d$ denotes the $\bm{i}$-th semantics in $\bm{\mathcal{D}_r}$.

\begin{equation}
\bm{\mathcal{D}^c_r} = \frac{1}{n} \sum_{\bm{i} = 1}^n \bm{\mathcal{D}^i_r}
\end{equation}

\subsection{Dictionary Lookup}
In this work, the composition operator $\phi$ is used to generate context vector $\bm{C^h_r} \in R^{d_c}$. In particular, we implement $\phi$ in the following four ways, where $\circ$ denotes Hadamard product and $\star$ means compressing the tensor product with circular correlation \cite{hole}. It should be noted that $d_c = 2d$ for $\phi_{concat}$, and $d_c = d$ for others, when the question is $(?,r,t)$, we replace $h$ with $t$.

\begin{equation}
\phi_{sum}(h, r) = \bm{h}+\bm{\mathcal{D}^c_r}
\end{equation}

\begin{equation}
\phi_{concat}(h, r) = [\bm{h};\bm{\mathcal{D}^c_r}]
\end{equation}

\begin{equation}
\phi_{mult}(h, r) = \bm{h} \circ \bm{\mathcal{D}^c_r}
\end{equation}

\begin{equation}
\phi_{corr}(h, r) = \bm{h} \star \bm{\mathcal{D}^c_r}
\end{equation}

After that, we can obtain the lookup vector $\bm{L^h_r} \in R^{n}$ via a simple MLP as follows, where $\bm{W_L} \in R^{d_c \times n}$ and $\bm{b_L} \in R^{n}$ are learnable weights and biases respectively, $\sigma$ represents the nonlinear activation function, the $softmax$ operator is applied to ensure that the sum of $\bm{L^h_r}$ equals to 1. Actually, $\bm{L^h_r}$ can be understood as the attention vector about $\bm{\mathcal{D}_r}$.

\begin{equation}
\bm{G^h_r} = MLP(\bm{C^h_r}) = \sigma(\bm{W_LC^h_r+b_L})
\end{equation}

\begin{equation}
\bm{L^h_r} = softmax(\bm{G^h_r}), \bm{L^{h_i}_r} = \frac{e^{\bm{G^{h_i}_r}}}{\sum_{j=1}^{n}e^{\bm{G^{h_j}_r}}}
\end{equation}

Ultimately, the fine-grained semantics $\bm{\mathcal{D}^h_r} \in R^d$ of relation \textit{r} under head entity \textit{h} is the product of lookup vector $\bm{L^h_r}$ and dictionary $\bm{\mathcal{D}_r}$ as follows, which can be regarded as the weighted sum of semantics in the dictionary.

\begin{equation}
\bm{\mathcal{D}^h_r} = \bm{L^h_r}\bm{\mathcal{D}_r}
\end{equation}

\subsection{Score \& Loss Function}
Inspired by \cite{dog}, we apply the sigmoid function on equation (\ref{etranse}, \ref{edist}) to enable TransE and DistMult to output probability within [0,1], so that they can better adapt to the binary cross entropy loss. The updated formula are as follows.

\begin{equation}
TransE: \psi(\bm{h},\bm{r},\bm{t}) = \sigma(||\bm{h}+\bm{r}-\bm{t}||)
\end{equation}

\begin{equation}
DistMult: \psi(\bm{h},\bm{r},\bm{t}) = \sigma((\bm{h}\circ\bm{r})^T\bm{t})
\end{equation}

Then $\bm{\mathcal{D}^c_r}$ and $\bm{\mathcal{D}^h_r}$ are substituted into above formulas to get $S^c$ and $S^f$, which correspond to the score calculated from the central and fine-grained semantics respectively. Suppose the label is T, the binary cross entropy loss for $S^c$ and $S^f$ are $\mathcal{L}^c$ and $\mathcal{L}^f$, the target of training is to optimize $\mathcal{L}$ as follows, where $\lambda$ is a trade-off parameter keeping the consistency of the two semantics, which should be set smaller than 1 to prevent them from being too close to be distinguished. Note that the well-trained models only apply $S^f$ for prediction.

\begin{equation}
S^c = \psi(\bm{h}, \bm{\mathcal{D}^c_r}, \bm{t}), S^f = \psi(\bm{h}, \bm{\mathcal{D}^h_r}, \bm{t}) 
\end{equation}

\begin{equation}
\mathcal{L}^c = Tlog(S^c)+(1-T)log(1-S^c)
\end{equation}

\begin{equation}
\mathcal{L}^f = Tlog(S^f)+(1-T)log(1-S^f)
\end{equation}

\begin{equation}
\mathcal{L} = \mathcal{L}^f+\lambda\mathcal{L}^c
\end{equation}

\subsection{Sparseness Of Lookup (SOL)}

To determine whether a dictionary lookup is valid, we observe the sparseness of lookup (SOL) while training. The sparser the lookup vector is, the better it can focus on the most appropriate fine-grained semantics. Intuitively, the sum of squares is a good indicator. Suppose $\mathcal{J}(\bm{L^h_r})$ is the sum of squares of $\bm{L^h_r}$, since each value of $\bm{L^h_r}$ lies between 0 and 1, it's easy to get the following inequality, where $\mathcal{J}(\bm{L^h_r})$ equals 1 if and only if $\bm{L^h_r}$ is one-hot.

\begin{equation}
\mathcal{J}(\bm{L^h_r}) = \sum_{i=1}^{n}(\bm{L^{h_i}_r})^2 \leq \sum_{i=1}^{n}\bm{L^{h_i}_r} = 1
\end{equation}

On the other hand, according to the Cauchy inequality, the minimum value of $\mathcal{J}(\bm{L^h_r})$ can be obtained when all elements are $\frac{1}{n}$ as shown below.

\begin{equation}
\mathcal{J}(\bm{L^h_r}) = \frac{1}{n}\sum_{i=1}^{n}n(\bm{L^{h_i}_r})^2 \geq \frac{1}{n}(\sum_{i=1}^{n}\bm{L^{h_i}_r})^2 = \frac{1}{n}
\end{equation}

Then we normalize $\mathcal{J}(\bm{L^h_r})$ as follows to get $SOL(\bm{L^h_r}) \in [0,1]$. Its value is 0 when $\bm{L^h_r}$ is uniform, the lookup is invalid in this case as $\bm{\mathcal{D}^h_r}$ always equals to $\bm{\mathcal{D}^c_r}$. When $SOL(\bm{L^h_r})$ is 1, there exists the greatest distinction between $\bm{\mathcal{D}^h_r}$ and $\bm{\mathcal{D}^c_r}$.

\begin{equation}
SOL(\bm{L^h_r}) = \frac{\mathcal{J}(\bm{L^h_r})-\frac{1}{n}}{1-\frac{1}{n}}
\end{equation}

\subsection{Dictionary Alignment Entity (DAE)}

Although the dictionary size of each relation maintains consistency globally, there exist differences in the diversity of fine-grained semantics among different relations indeed, and such differences can be synchronously reflected in their corresponding entities. For example, the relation \textit{PartOf} has three fine-grained semantics in Fig.~\ref{example}(a), the head entities connecting with \textit{PartOf} can also be divided into three classes including \textit{Location}, \textit{Object} and \textit{Concept}. While for relation \textit{PresidentOf}, the semantics is unique, and its head entities can all be classified as \textit{Individual}. An accurate dictionary lookup is expected to align the diversity of both fine-grained semantics and the relation's corresponding entities.

Considering that both dictionary and relation-connected entities can be treated as a set containing multiple vectors, we define $DIV$ metric based on cosine similarity to measure their internal diversity. Specifically, given vector set $V$, we define $V_c = \frac{1}{|V|} \sum_{v \in V} v$ as its mean vector, after obtaining the average of cosine similarity between all $v$ and $V_c$, we transform it as follows to get $DIV (V)$ ranging from 0 to 1, 0 means that vectors are the same, while 1 means that they are opposite.

\begin{equation}
DIV(V) = \frac{1-\frac{1}{|V|} \sum_{v \in V} \frac{v \cdot V_c}{|v| \times |V_c|}}{2}
\end{equation}

On the basis of above definitions, we propose the dictionary alignment entity (DAE) to evaluate the accuracy of dictionary lookup. Concretely, given relation \textit{r}, suppose $\bm{\mathcal{H}_r}\!\!=\!\!\{\bm{h}\!\mid\!(h, r, t) \in \mathcal{G}\}$ is the set of all entity embeddings connecting it (note that when the question is $(?,r,t)$, $\bm{\mathcal{H}_r}\!\!=\!\!\{\bm{t}\!\mid\!(h, r, t) \in \mathcal{G}\}$), then $DAE(r)$ is the absolute difference between $DIV(\bm{\mathcal{H}_r})$ and $DIV(\bm{\mathcal{D}_r})$ as follows, it's obvious that a smaller DAE corresponds to a stronger alignment.

\begin{equation}
DAE(r) = \vert DIV(\bm{\mathcal{H}_r})-DIV(\bm{\mathcal{D}_r}) \vert
\end{equation}

\section{Experiments}

In this section, we will extend several representative models with CoDLR, the experimental setting and results analysis will be introduced below.

\subsection{Experimental Setting}

\paragraph{Datasets}

Several large-scale knowledge graphs including Freebase \cite{freebase}, WordNet \cite{wordnet}, NELL \cite{nell} and YAGO3 \cite{yago3} have been built to store facts from real world, in this work, we adopt FB15k-237 \cite{fb15k237} and WN18RR \cite{conve} to confirm the performance of CoDLR. Note that to resolve the test leakage problem, the two benchmarks have their inverse relations removed. In particular, FB15k-237 provides general facts including \textit{countries}, \textit{companies} and \textit{individuals}, while WN18RR contains semantic knowledge of lexicons such as \textit{synonym} and \textit{hyponym}, more details are shown in Table \ref{dataset}.

\paragraph{Metrics}

\begin{table}[!t]
	\caption{Statistics of datasets.}
	\setlength\tabcolsep{12pt}
	\renewcommand\arraystretch{0.84} 
	\centering
	\begin{tabular}{l|cc}
		\toprule
		\textbf{\textbf{Statistics}}                                  & \textbf{FB15k-237} & \textbf{WN18RR} \\ 
		\midrule
		Entities  & 14,541    & 40,943 \\ 
		Relations   & 237       & 11     \\ 
		Train Set	 & 272,115   & 86,835 \\ 
		Valid Set   & 17,535    & 3,034  \\ 
		Test Set       & 20,466    & 3,134  \\ 
		\bottomrule
	\end{tabular}
	\label{dataset}
\end{table}

\begin{table*}[!t]
\caption{The best hyperparameters we adopt for extended models on FB15k237.}
\centering
\begin{tabular}{l|ccccccc}
	\toprule
	Model          & $d$ & $\phi$        & $n$ & $\lambda$ & batch size & learning rate & epoch \\ \midrule
	CoDLR-TransE   & 100 & $\phi_{sum}$  & 7   & 0.001     & 256        & 0.001         & 400   \\
	CoDLR-DistMult & 200 & $\phi_{corr}$ & 5   & 0.1       & 1024       & 0.00015       & 400   \\
	CoDLR-ConvE    & 200 & $\phi_{mult}$ & 5   & 0.03      & 1024       & 0.00005       & 400  \\ \bottomrule
\end{tabular}
\label{hfb}
\end{table*}

\begin{table*}[!t]
\caption{The best hyperparameters we adopt for extended models on WN18RR.}
\centering
\begin{tabular}{l|ccccccc}
	\toprule
	Model          & $d$ & $\phi$        & $n$ & $\lambda$ & batch size & learning rate & epoch \\ \midrule
	CoDLR-TransE   & 100 & $\phi_{corr}$  & 5   & 0.01     & 128        & 0.00018         & 400   \\
	CoDLR-DistMult & 500 & $\phi_{mult}$ & 3   & 0.000005      & 16       & 0.0002       & 120   \\
	CoDLR-ConvE    & 200 & $\phi_{corr}$ & 4   & 0.6      & 1024       & 0.001       & 400  \\ \bottomrule
\end{tabular}
\label{hwn}
\end{table*}

To evaluate the performance of extended models on the link prediction task, we report commonly used ranking-based metrics including mean rank (MR), mean reciprocal rank (MRR), and Hits@k (H@k) for k = 1, 3, 10. Specifically, suppose $r_i$ is the ranking of the target tail entity for $i$-th triple, MR denotes the mean value of $r_i$, MRR is defined as the mean value of $\frac{1}{r_i}$ \cite{mrr}, while H@k measures the average proportion of $r_i$ less than k. To sum up, the better the model performs, the higher both MRR and H@k are, the lower MR is. To avoid the unfairness caused by the model's tendency to output same scores for all triples to be predicted, the random evaluation protocols are adopted \cite{reevaluation}. Besides, we filter out valid candidates different from the answer while training following \cite{transe}. 

\paragraph{Baselines}

To prove the effectiveness of CoDLR, we combine it with three expanded baselines, the results are compared with several contrast baselines, they are listed as follows.

\begin{itemize}
\item \textbf{Expanded baselines:} The additive model TransE \cite{transe}, the multiplicative model DistMult \cite{distmult}, and the neural network based model ConvE \cite{conve} are extended with CoDLR.
\item \textbf{Contrast baselines:} Additive models including TransH \cite{transh}, CTransR \cite{transr}, TransD \cite{transd}. The multiplicative model Complex \cite{complex}. Neural network based models including ConvKB \cite{convkb} and R-GCN \cite{rgcn}.
\end{itemize}

\begin{table*}[!t]
	\caption{Link prediction results for FB15k-237 and WN18RR. We implement TransE, DistMult, and ConvE as well as their extended variants, while the performance of other models is taken from original or latest published papers. }
	\setlength\tabcolsep{5.5pt}
	\renewcommand\arraystretch{0.92} 
	\centering
	\begin{tabular}{l|ccccc|ccccc}
		\toprule
		\multirow{2}{*}{Model} & \multicolumn{5}{c|}{FB15k-237}                                                                                 & \multicolumn{5}{c}{WN18RR}                                                                                      \\
		& \textbf{MR}        & \textbf{MRR}         & \textbf{H@10}        & \textbf{H@3}         & \textbf{H@1}         & \textbf{MR}         & \textbf{MRR}         & \textbf{H@10}        & \textbf{H@3}         & \textbf{H@1}         \\ \midrule
		TransH                 & 311                & 0.211                & 0.386                & 0.224                & 0.132                & -                   & -                    & -                    & -                    & -                    \\
		CTransR                & 279                & \textbf{0.298}                & \textbf{0.469}                & 0.301                & \textbf{0.198}                & -                   & -                    & -                    & -                    & -                    \\
		TransD                 & \textbf{256}                & 0.286                & 0.453                & 0.291                & 0.179                & -                   & -                    & -                    & -                    & -                    \\
		Complex                & -                  & 0.247                & 0.428                & 0.275                & 0.158                & -                   & \textbf{0.440}                & 0.510                & \textbf{0.460  }              & \textbf{0.410}                \\
		ConvKB                 & 309                & 0.243                & 0.421                & \textbf{0.371}                & 0.155                & \textbf{3433}                & 0.249                & \textbf{0.524}                & 0.417                & 0.057                \\
		R-GCN                  & -                  & 0.248                & 0.417                & -                    & 0.151                & -                   & -                    & -                    & -                    & -                    \\ \midrule
		TransE                 & 192                & 0.334                & 0.513                & 0.368                & 0.244                & 3234                & 0.228                & 0.517                & 0.348                & 0.064                \\
		CoDLR-TransE           & {\ul \textbf{185}} & {\ul \textbf{0.340}} & {\ul \textbf{0.517}}       & {\ul \textbf{0.375}} & {\ul \textbf{0.250}} & {\ul \textbf{3129}} & \textbf{0.357}       & {\ul \textbf{0.535}} & \textbf{0.417}       & \textbf{0.258}       \\ \midrule
		DistMult               & 337                & 0.286                & 0.440                & 0.311                & 0.209                & 7541                & 0.416                & 0.483                & 0.433                & 0.383                \\
		CoDLR-DistMult         & \textbf{263}       & \textbf{0.300}       & \textbf{0.464}       & \textbf{0.327}       & \textbf{0.218}       & \textbf{5925}       & \textbf{0.438}       & \textbf{0.494}       & \textbf{0.452}       & \textbf{0.407}       \\ \midrule
		ConvE                  & 209                & 0.333                & 0.510                & 0.366                & 0.244                & 4797                & 0.435                & 0.480                & 0.443                & 0.411                \\
		CoDLR-ConvE            & \textbf{201}       & \textbf{0.338}       & \textbf{0.517} & \textbf{0.373}       & \textbf{0.248}       & \textbf{4723}       & {\ul \textbf{0.456}} & \textbf{0.520}       & {\ul \textbf{0.467}} & {\ul \textbf{0.425}} \\ \bottomrule
	\end{tabular}
	\label{compare_result}
\end{table*}

\paragraph{Implementations}
We implement CoDLR in PyTorch \cite{pytorch}, the dictionaries, embeddings, and other model parameters are initialized randomly, the Adam optimizer \cite{adam} is applied to speed up the training process. Inspired by  \cite{dog}, we use \textit{kvsAll} to train models, where all triples are labeled as either positive or negative according to whether they appear in the train set or not. In this work, hyperparameters are confirmed by grid search, the optimal ones are determined by the best MRR evaluated on the valid set. To be more specific, the composition operator $\phi$ can be selected from four implementations, the dictionary size $n$ ranges from 2 to 10, the dimension $d$ is searched in $\{100, 200, 300, 400\}$, the trade-off parameter $\lambda$ is within the range of 0 to 1, the learning rate varies from $1e-4$ to $1e-2$, the batch size is tuned amongst $\{16, 32, \dots, 1024\}$. To ensure the reproducibility, the hyperparameters we adopt for extended models on FB15k237 and WN18RR are shown in Table \ref{hfb} and Table \ref{hwn} respectively, it can be found that the difference of models and datasets does affect the choice of hyperparameters.

\subsection{Performance Comparison}

We evaluate the performance of CoDLR-TransE, CoDLR-DistMult, and CoDLR-ConvE in Table \ref{compare_result}, the results are divided into four groups, note that result in bold means the best in a group, while result underlined indicates the best across groups, and "-" denotes missing results. It's obvious that the extended models surpass the original models in all datasets and metrics, verifying that taking the fine-grained semantics into consideration while dealing with link prediction tasks is valuable. Especially, CoDLR-TransE achieves 3.65\% MR and 2.46\% H@1 performance gain on FB15k237, 53.58\% MRR, 19.83\% H@3, and 303.13\% H@1 performance gain on WN18RR, CoDLR-ConvE achieves 4.83\% MRR, 8.33\% H@10, 5.42\% H@3 and 3.41\% H@1 performance gain on WN18RR, enabling them to surpass other baselines significantly over the two benchmarks.

To further explore how our proposed method works, we calculate the average improvements of metrics as 9.28\% MR, 12.48\% MRR, 3.62\% H@10, 6.43\% H@3, and 53.53\% H@1. It can be summarized that the lower the value of k, the greater the improvement on H@k. Actually, a smaller k corresponds to a higher requirement for ranking multiple indistinguishable candidates sharing similar semantics, which indicates that fine-grained semantics introduced by CoDLR brings more refined predictions.

In addition, compared with CTransR, the conventional two-stage fine-grained semantics model based on clustering algorithm, the aforementioned extended models perform significantly better, proving the effectiveness of our end-to-end schema.

\subsection{Ablation Study}

\begin{table*}[!t]
	\caption{Results of ablation study for different modules on FB15k-237 and WN18RR.}
	\setlength\tabcolsep{5.5pt}
	\renewcommand\arraystretch{0.98}
	\centering
	\begin{tabular}{l|ccccc|ccccc}
		\toprule
		\multirow{2}{*}{Model} & \multicolumn{5}{c|}{FB15k-237}                                                                                 & \multicolumn{5}{c}{WN18RR}                                                                                      \\
		& \textbf{MR}        & \textbf{MRR}         & \textbf{H@10}        & \textbf{H@3}         & \textbf{H@1}         & \textbf{MR}         & \textbf{MRR}         & \textbf{H@10}        & \textbf{H@3}         & \textbf{H@1}         \\ \midrule
		CoDLR-TransE           & 185                & {\ul \textbf{0.340}} & 0.517                & {\ul \textbf{0.375}} & {\ul \textbf{0.250}} & 3129                & \textbf{0.357}       & {\ul \textbf{0.535}} & \textbf{0.417}       & \textbf{0.258}       \\ 
		CSDLR-TransE           & 191                & 0.334                & 0.510                & 0.369                & 0.243                & 2997                & 0.235                & 0.531                & 0.362                & 0.064                \\
		REDLR-TransE           & 185                & 0.338                & 0.517                & 0.373                & 0.247                & {\ul \textbf{2981}} & 0.241                & 0.534                & 0.362                & 0.076                \\
		FSCoDLR-TransE         & {\ul \textbf{183}} & 0.339                & {\ul \textbf{0.522}} & 0.375                & 0.247                & 3165                & 0.348                & 0.529                & 0.416                & 0.244                \\ \midrule
		CoDLR-DistMult         & \textbf{263}       & \textbf{0.300}       & \textbf{0.464}       & \textbf{0.327}       & \textbf{0.218}       & \textbf{5925}       & \textbf{0.438}       & \textbf{0.494}       & \textbf{0.452}       & \textbf{0.407}       \\
		CSDLR-DistMult         & 378                & 0.284                & 0.435                & 0.310                & 0.208                & 6282                & 0.426                & 0.490                & 0.442                & 0.393                \\
		REDLR-DistMult         & 282                & 0.295                & 0.457                & 0.321                & 0.214                & 6956                & 0.433                & 0.490                & 0.449                & 0.402                \\
		FSCoDLR-DistMult       & 272                & 0.297                & 0.461                & 0.323                & 0.216                & 6989                & 0.431                & 0.486                & 0.448                & 0.402                \\ \midrule
		CoDLR-ConvE            & 201                & \textbf{0.338}       & {\ul \textbf{0.517}} & \textbf{0.373}       & \textbf{0.248}       & 4723                & {\ul \textbf{0.456}} & \textbf{0.520}       & {\ul \textbf{0.467}} & {\ul \textbf{0.425}} \\
		CSDLR-ConvE            & 204                & 0.336                & 0.517                & 0.369                & 0.246                & 4718                & 0.454                & 0.518                & 0.464                & 0.422                \\
		REDLR-ConvE            & \textbf{196}       & 0.336                & 0.515                & 0.370                & 0.246                & 4870                & 0.456                & 0.518                & 0.466                & 0.424                \\
		FSCoDLR-ConvE          & 203                & 0.335                & 0.515                & 0.368                & 0.245                & \textbf{4643}       & 0.450                & 0.507                & 0.460                & 0.420   \\ \bottomrule             
	\end{tabular}
	\label{ablation} 
\end{table*}

\begin{figure*}[!t]
	\centering
	\subfloat[]{\includegraphics[width=0.33\hsize]{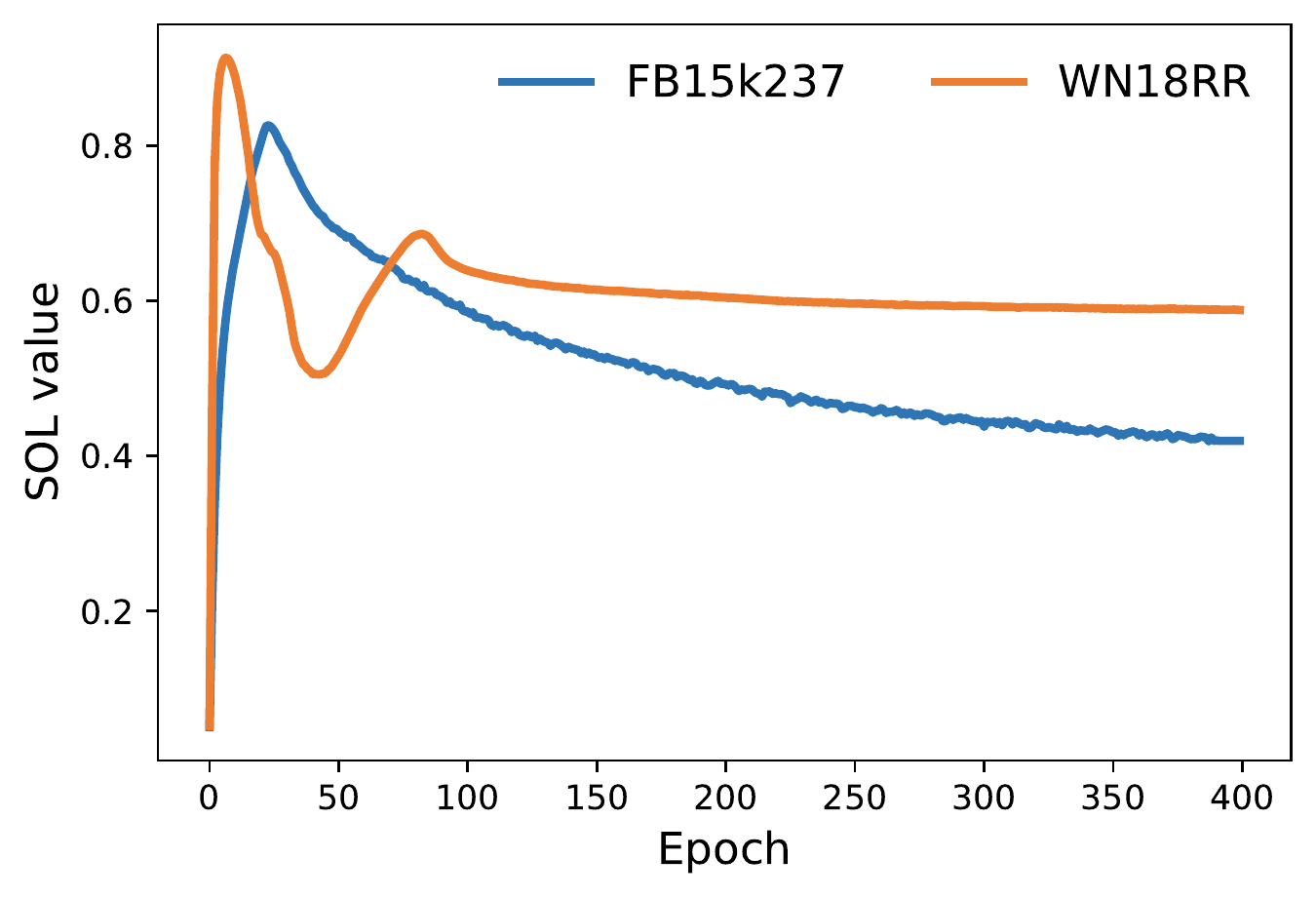}}
	\subfloat[]{\includegraphics[width=0.33\hsize]{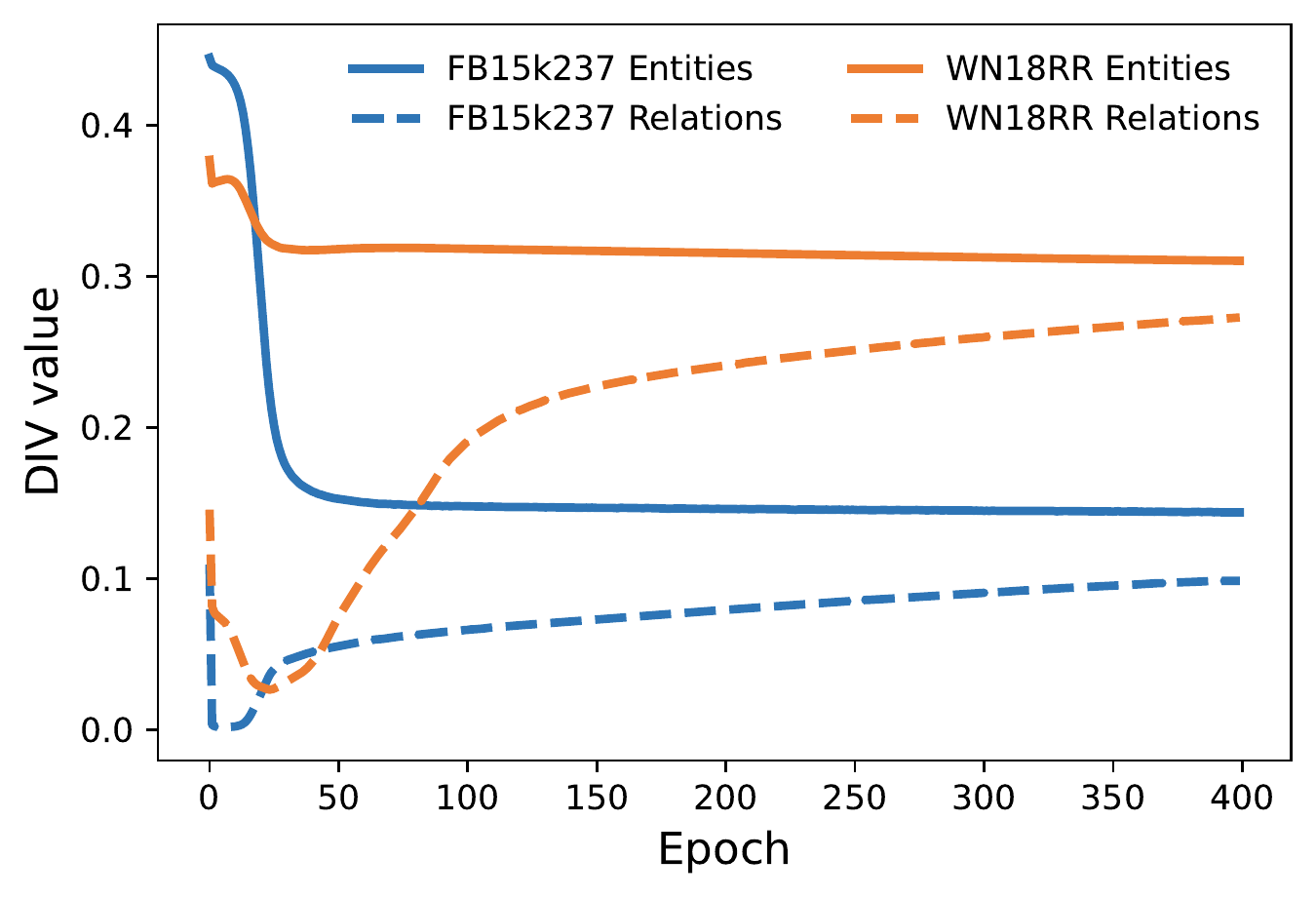}}
	\subfloat[]{\includegraphics[width=0.335\hsize]{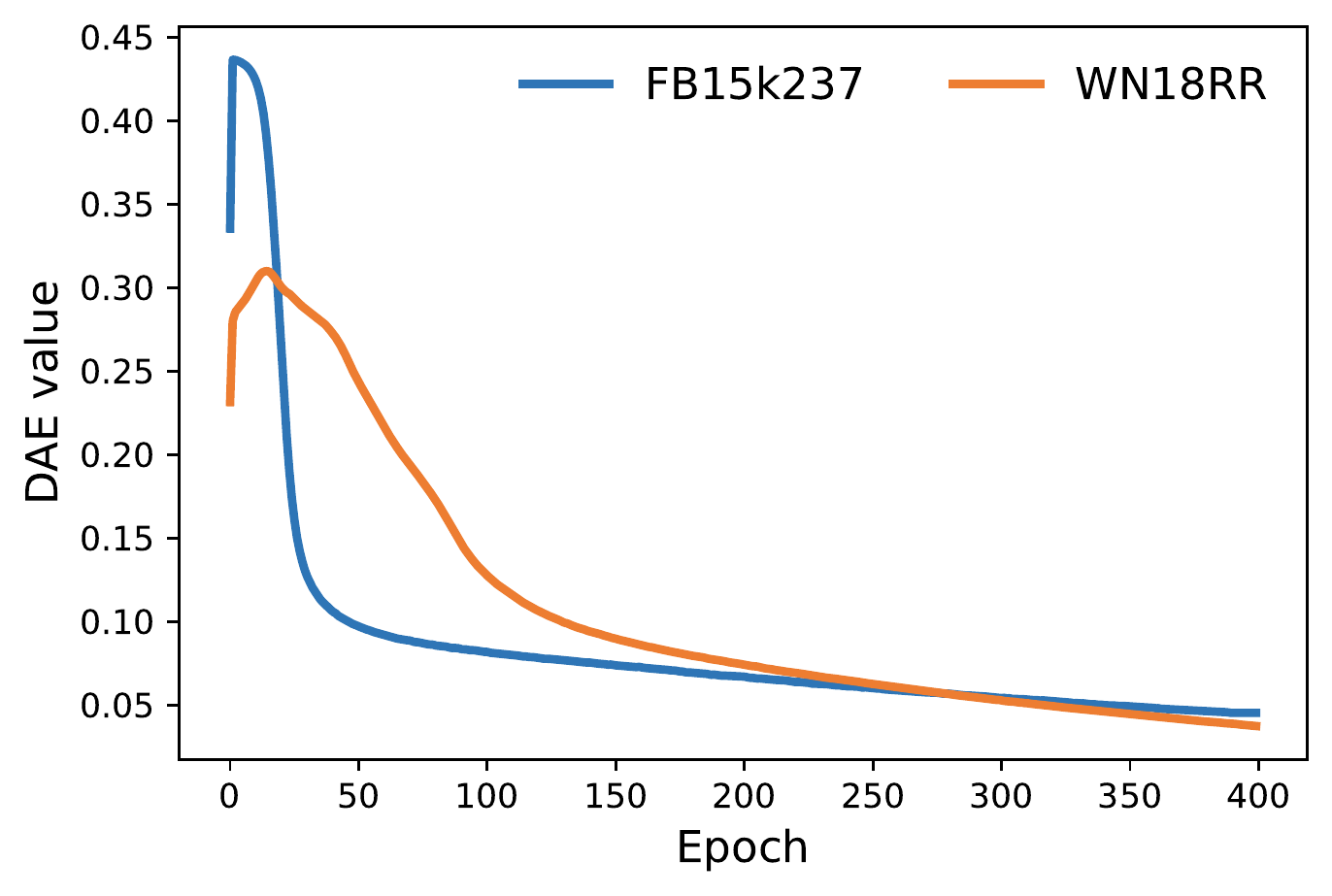}}
	\caption{Training curves of different metrics for CoDLR-TransE on FB15k237 and WN18RR. (a) The SOL curves for lookup vectors learned by CoDLR-TransE. (b) The DIV curves for relation-connected entities and relation dictionaries learned by CoDLR-TransE. (c) The DAE curves for relation-connected entities and relation dictionaries learned by CoDLR-TransE.}
	\label{metrics}
\end{figure*}

\begin{figure*}[!t]
\centering
\subfloat[]{\includegraphics[width=0.33\hsize]{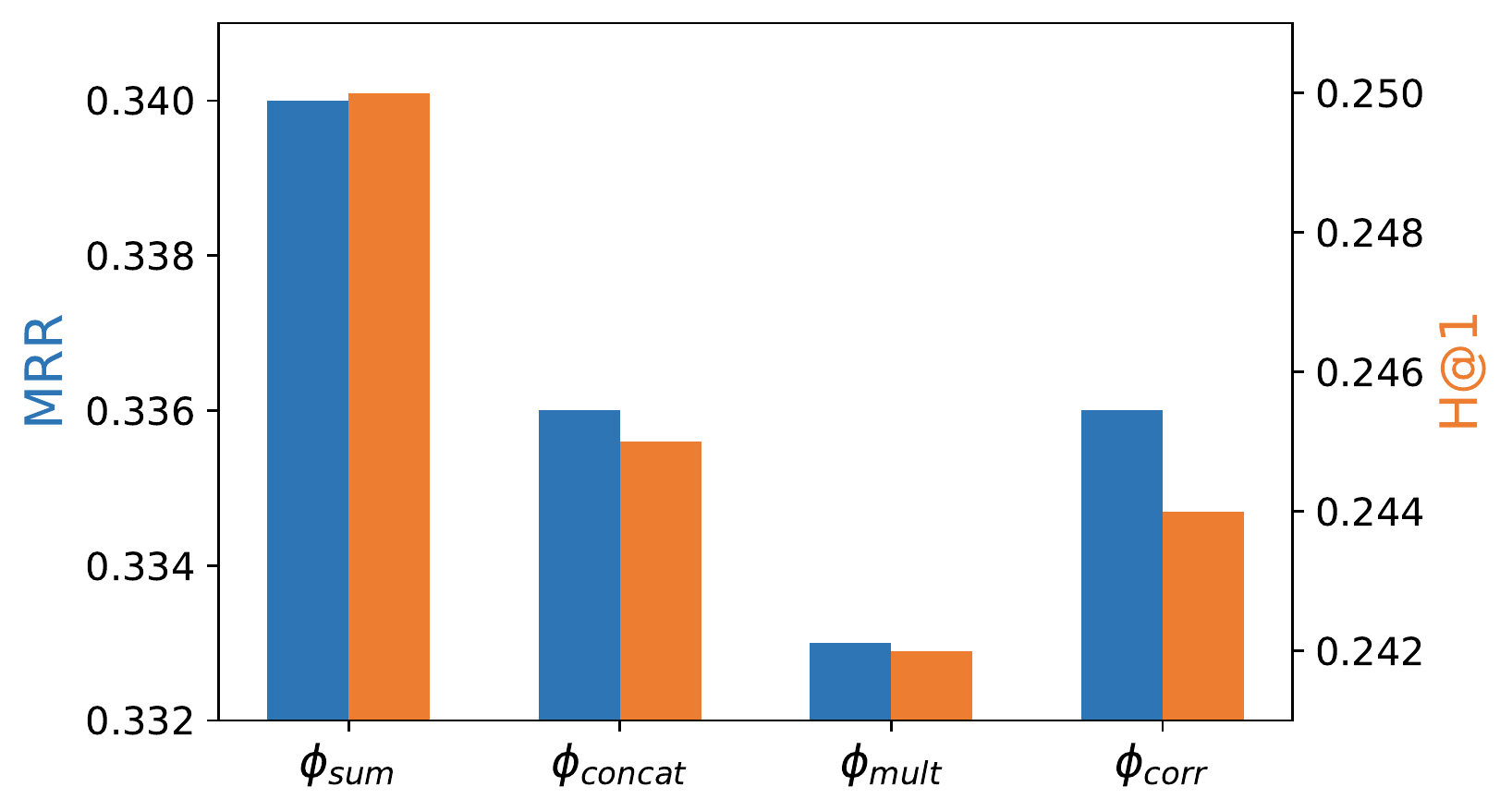}}
\subfloat[]{\includegraphics[width=0.33\hsize]{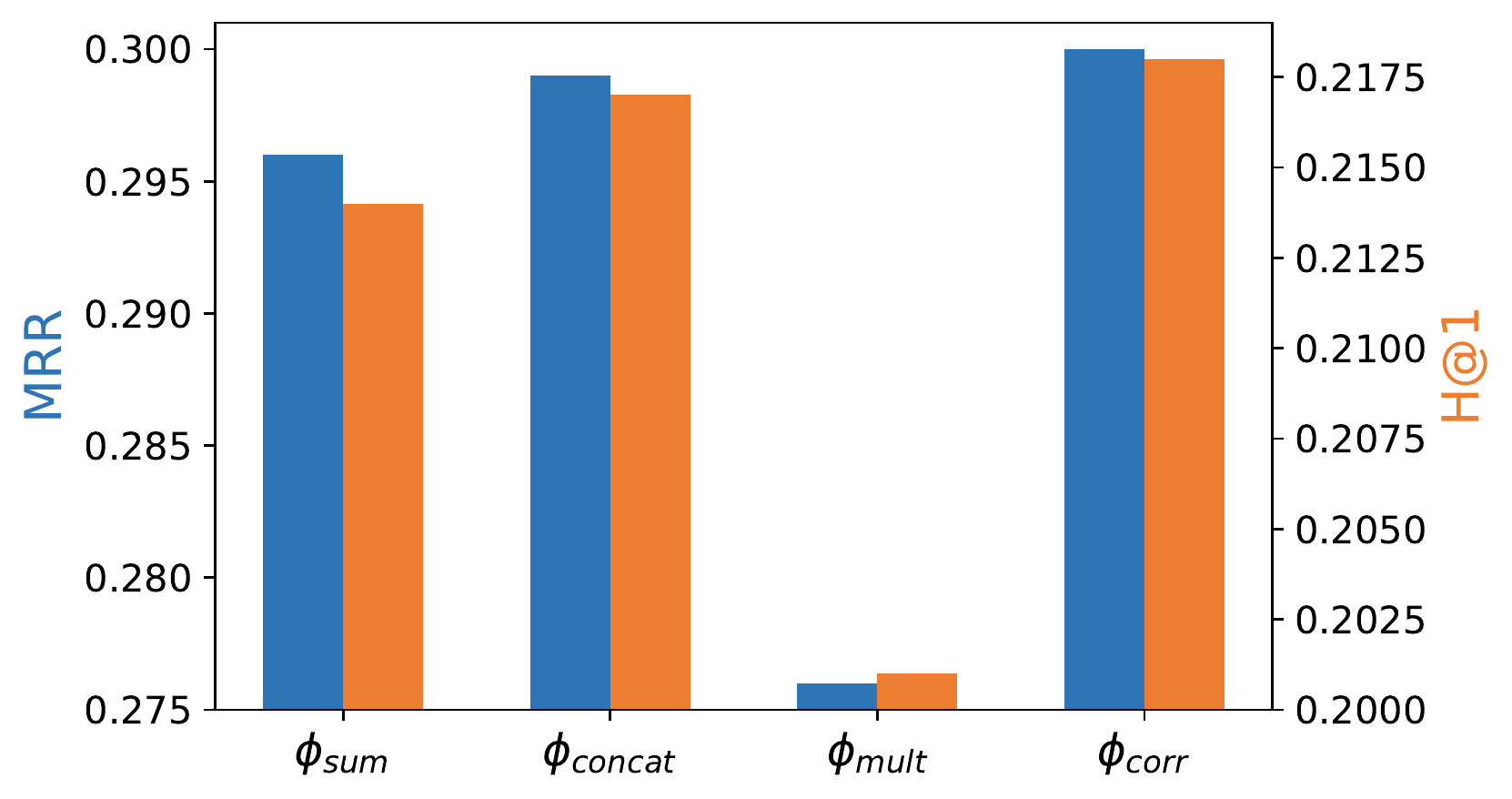}}
\subfloat[]{\includegraphics[width=0.33\hsize]{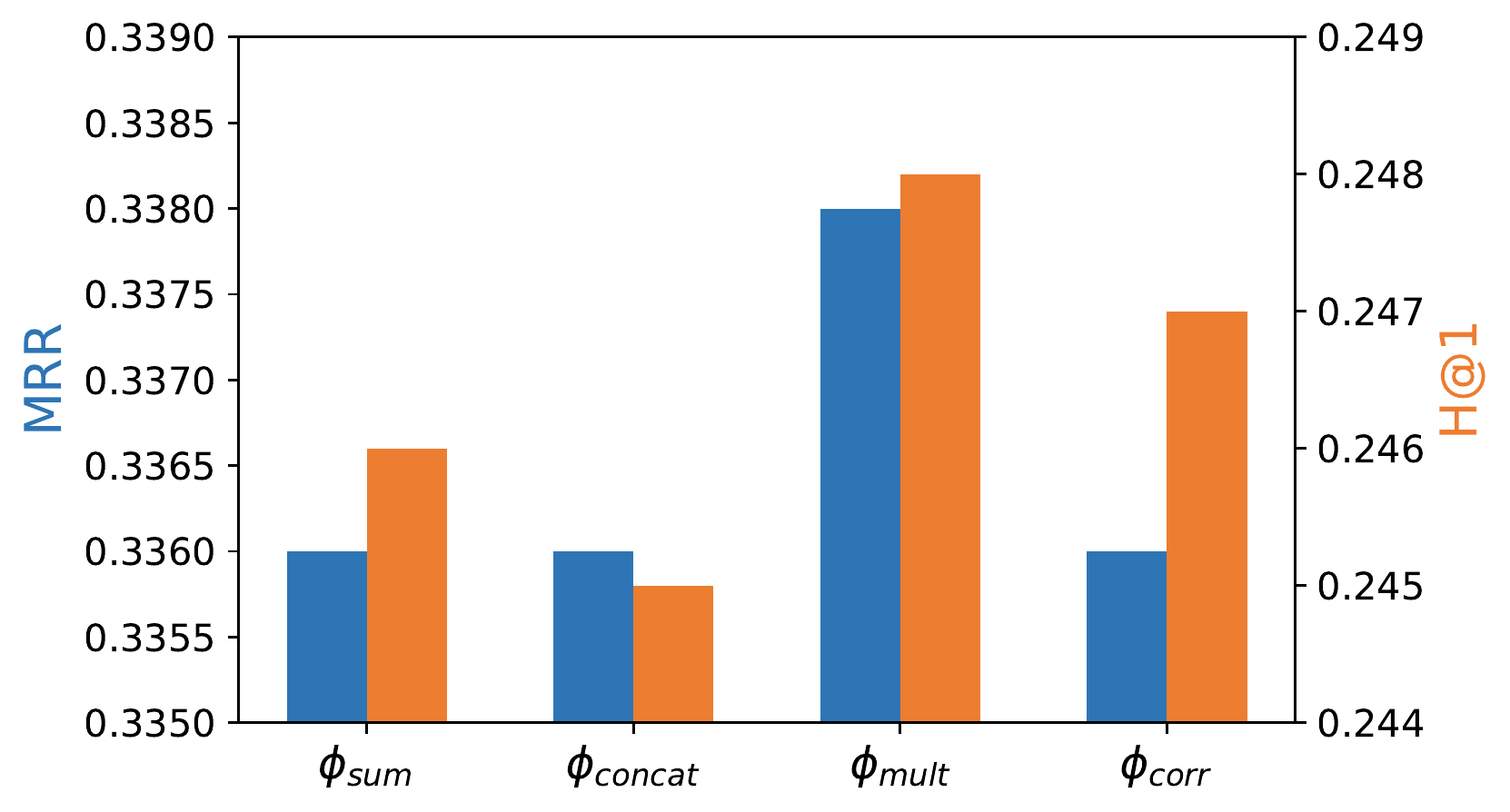}}

\subfloat[]{\includegraphics[width=0.33\hsize]{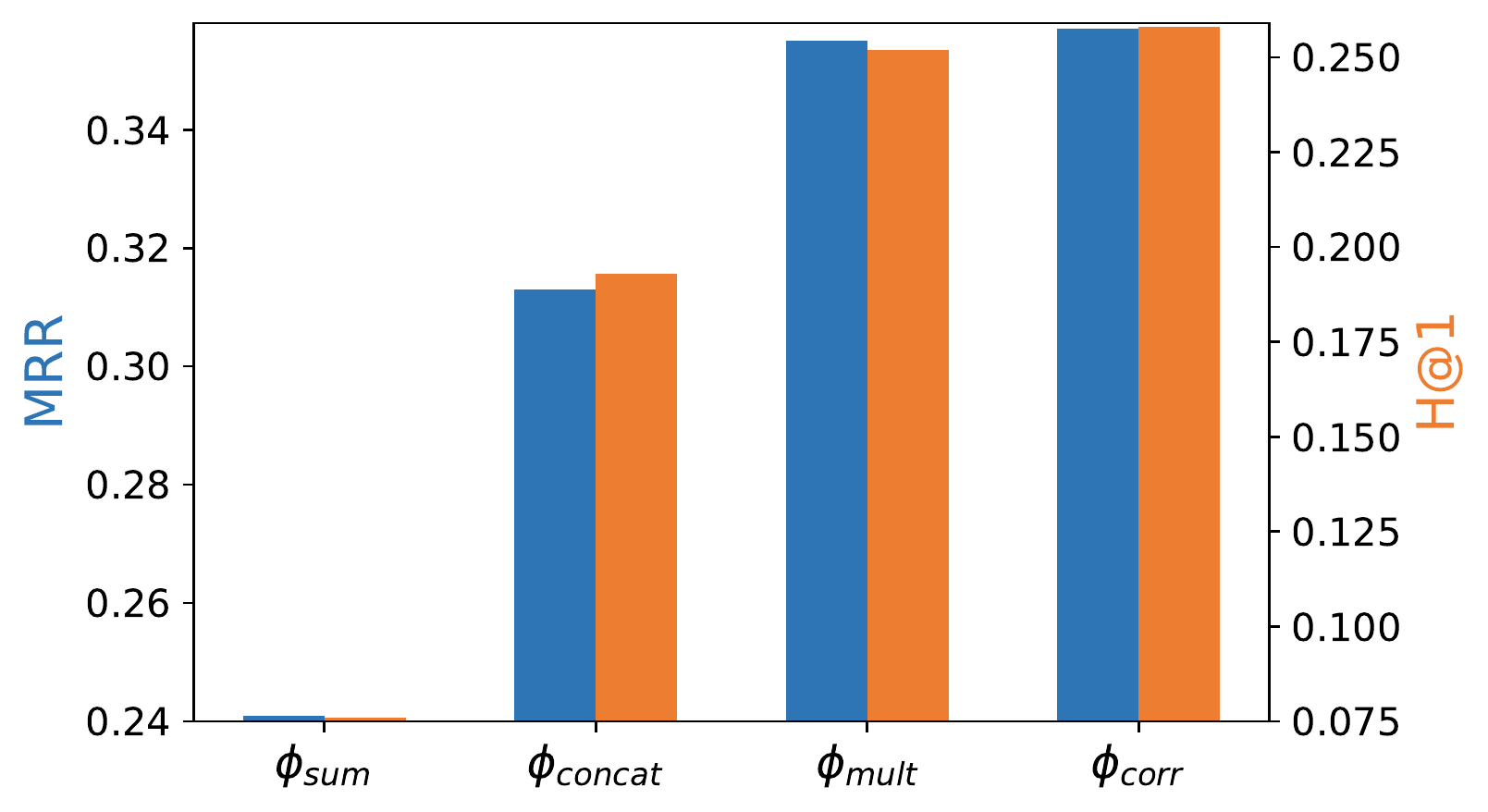}}
\subfloat[]{\includegraphics[width=0.33\hsize]{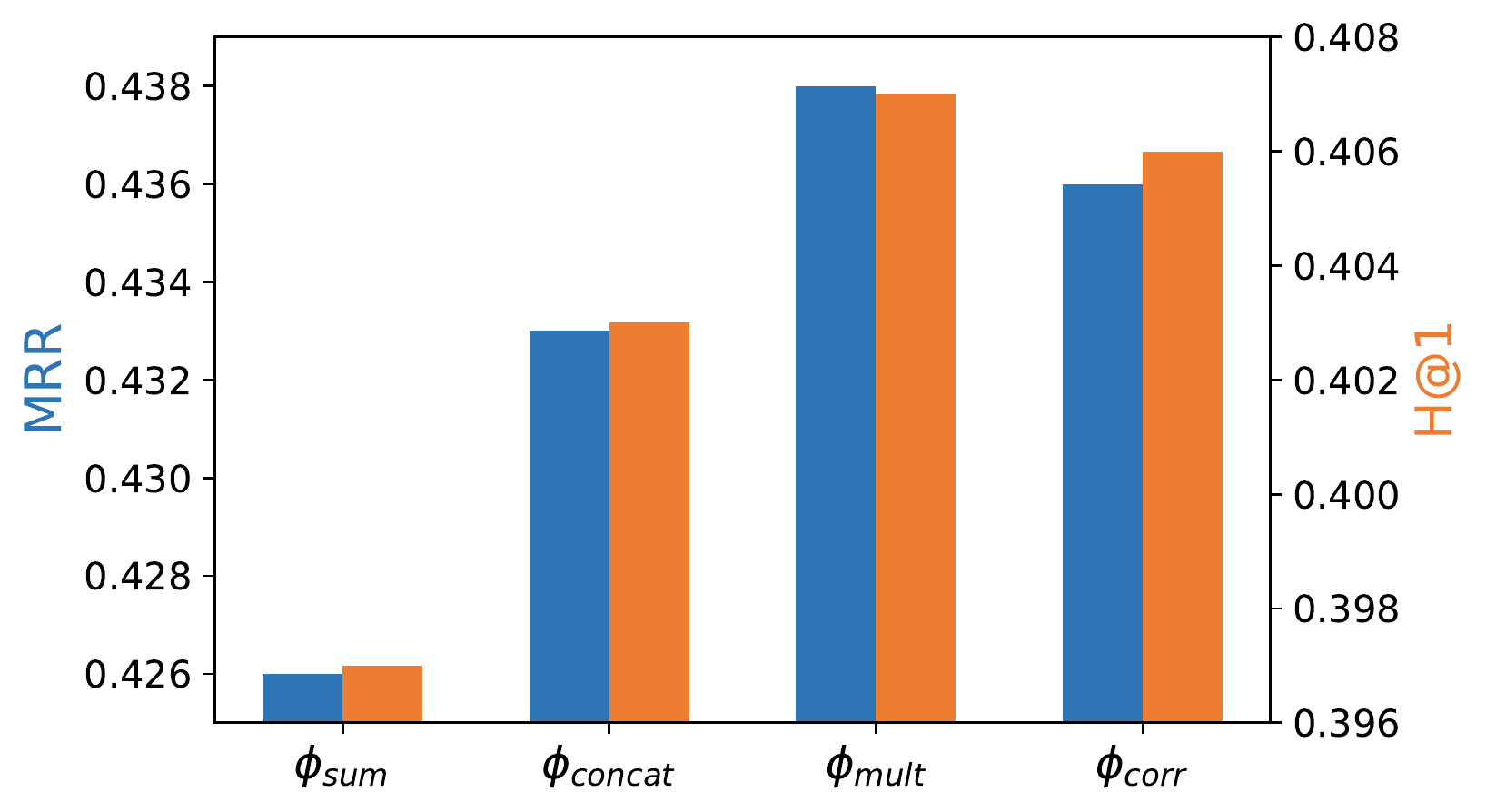}}
\subfloat[]{\includegraphics[width=0.33\hsize]{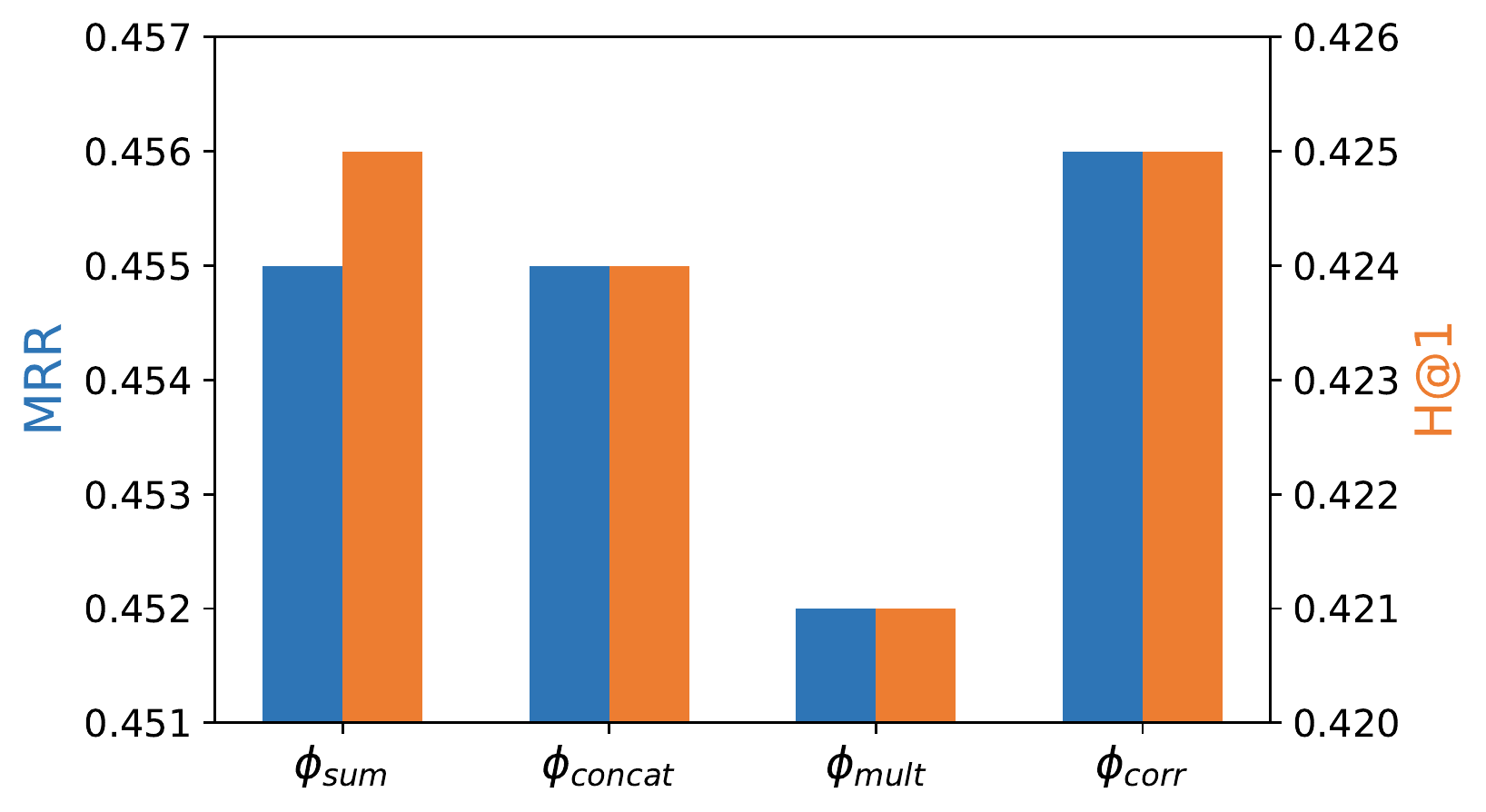}}
\caption{The influence of $\phi$ on the performance of different models. (a) The CoDLR-TransE on FB15k237. (b) The CoDLR-DistMult on FB15k237. (c) The CoDLR-ConvE on FB15k237. (d) The CoDLR-TransE on WN18RR. (e) The CoDLR-DistMult on WN18RR. (f) The CoDLR-ConvE on WN18RR.}
\label{contex}
\end{figure*}

\begin{figure*}[!t]
\centering
\subfloat[]{\includegraphics[width=0.33\hsize]{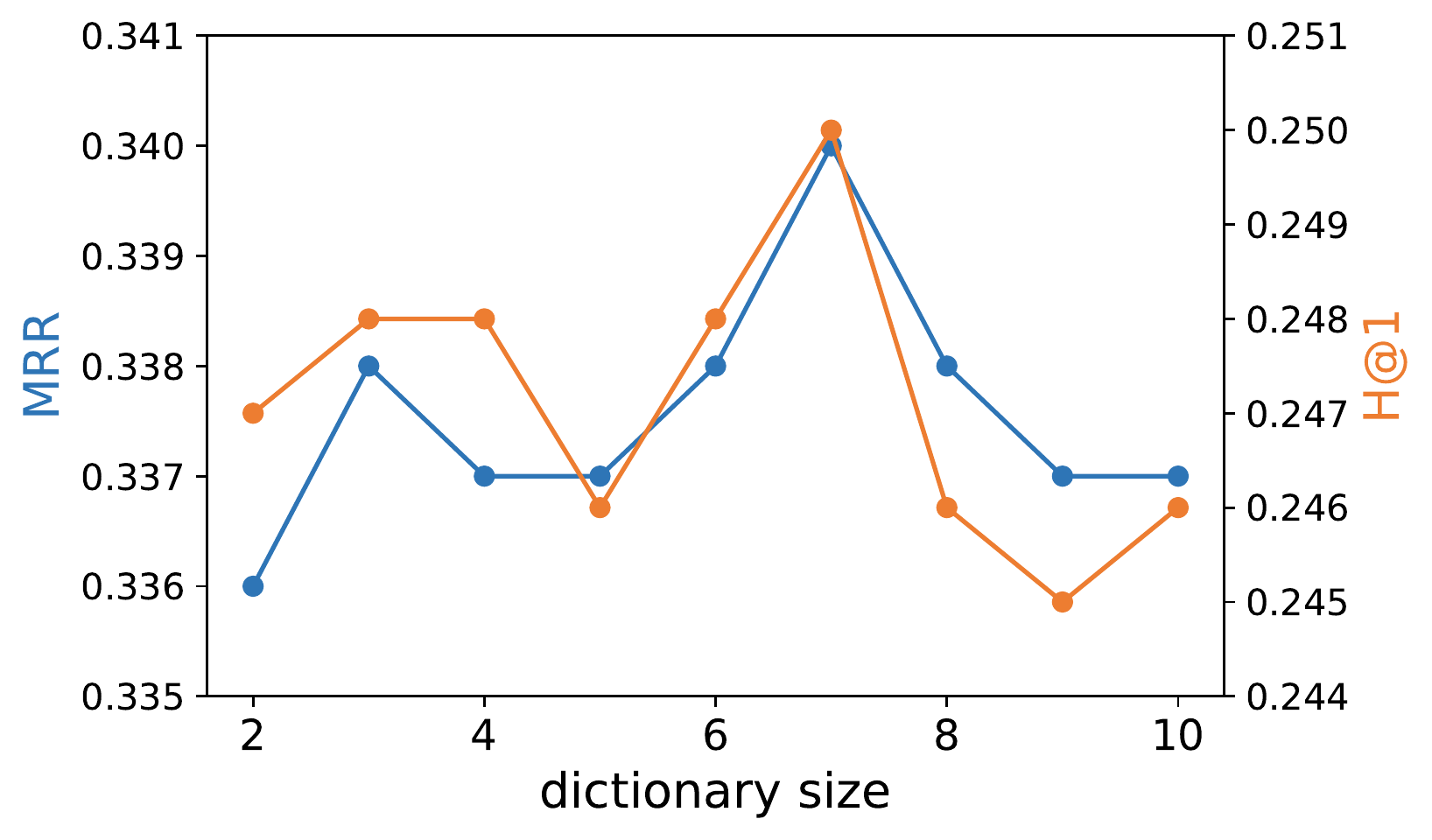}}
\subfloat[]{\includegraphics[width=0.33\hsize]{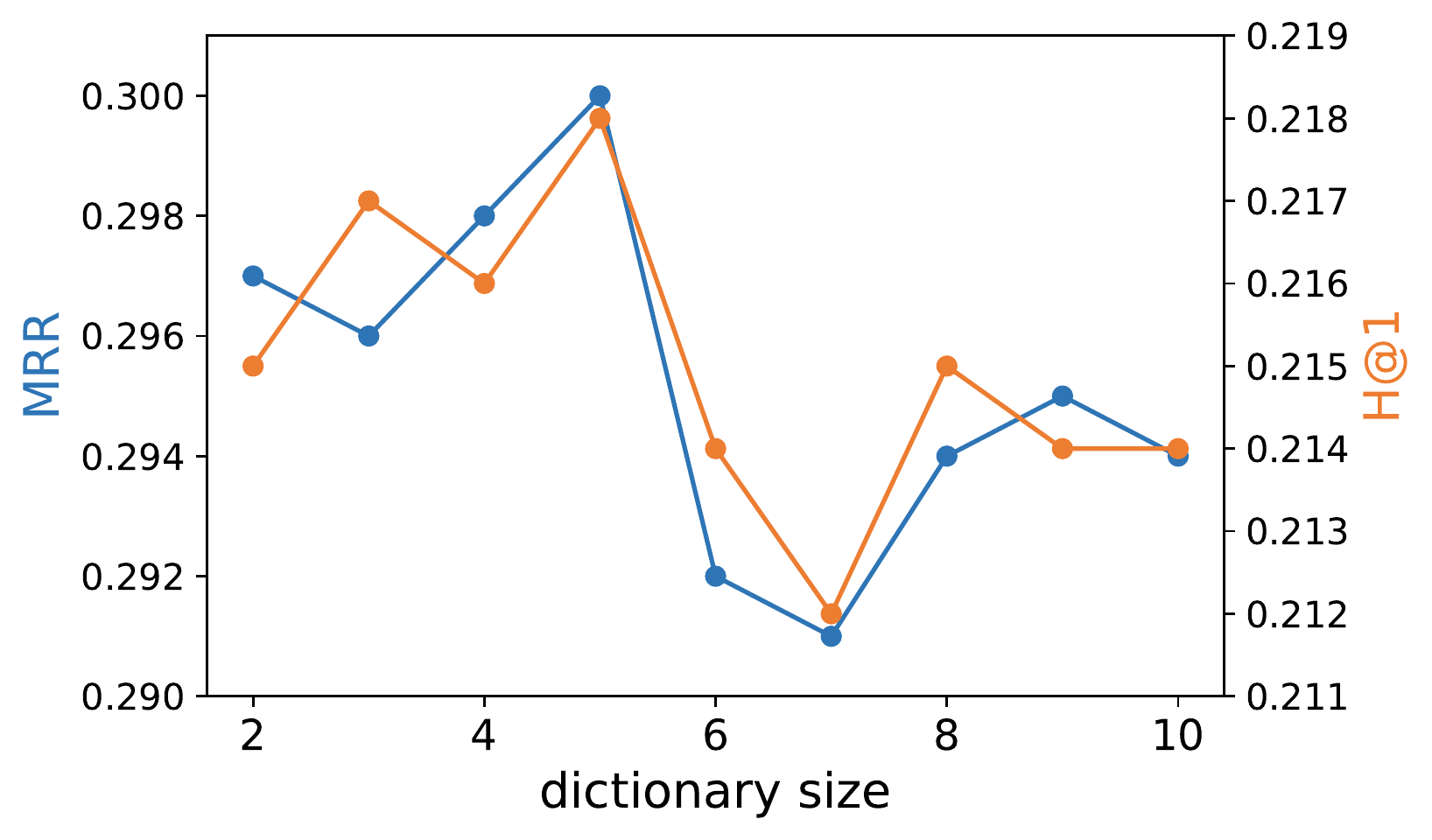}}
\subfloat[]{\includegraphics[width=0.33\hsize]{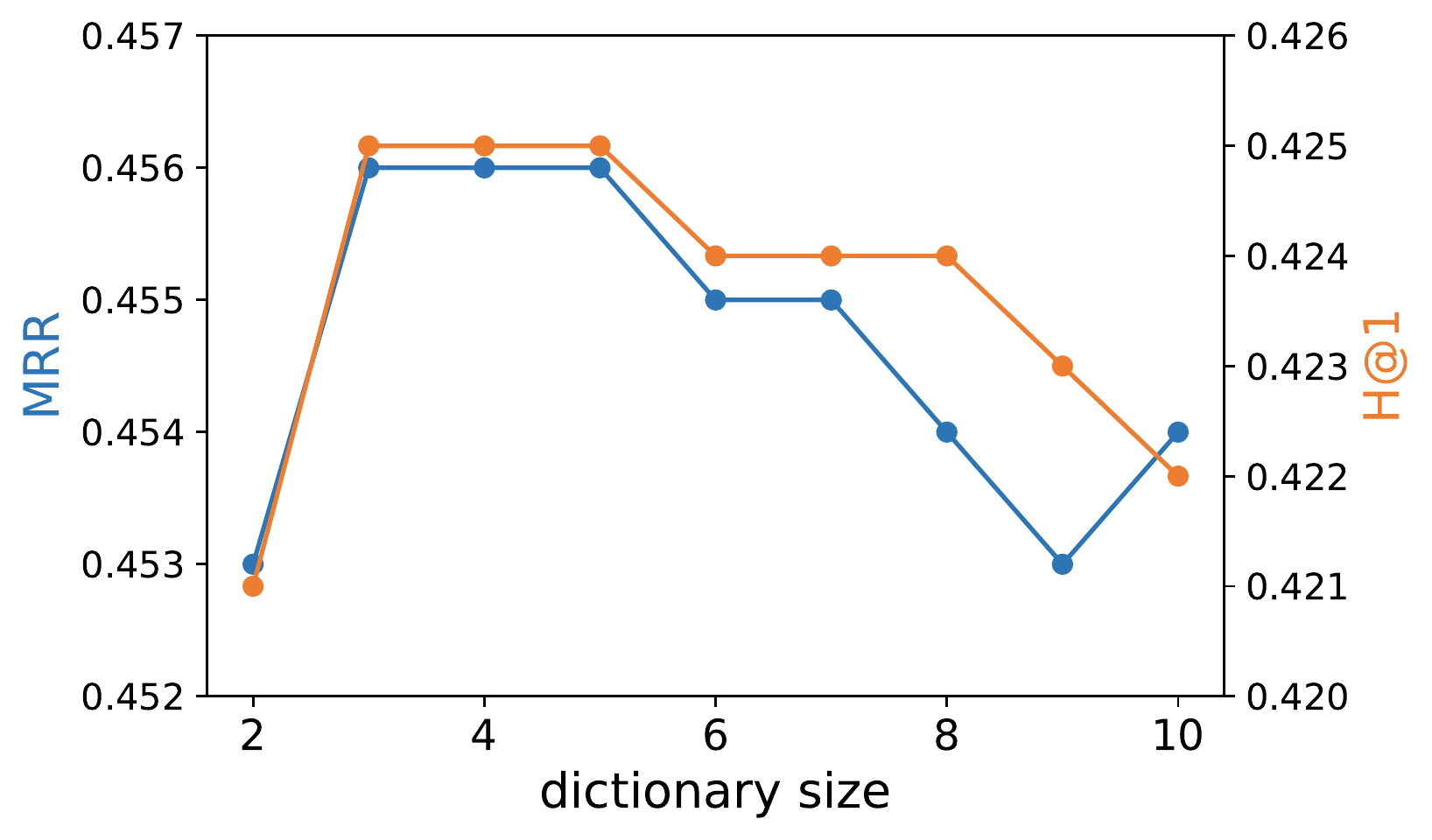}}

\subfloat[]{\includegraphics[width=0.33\hsize]{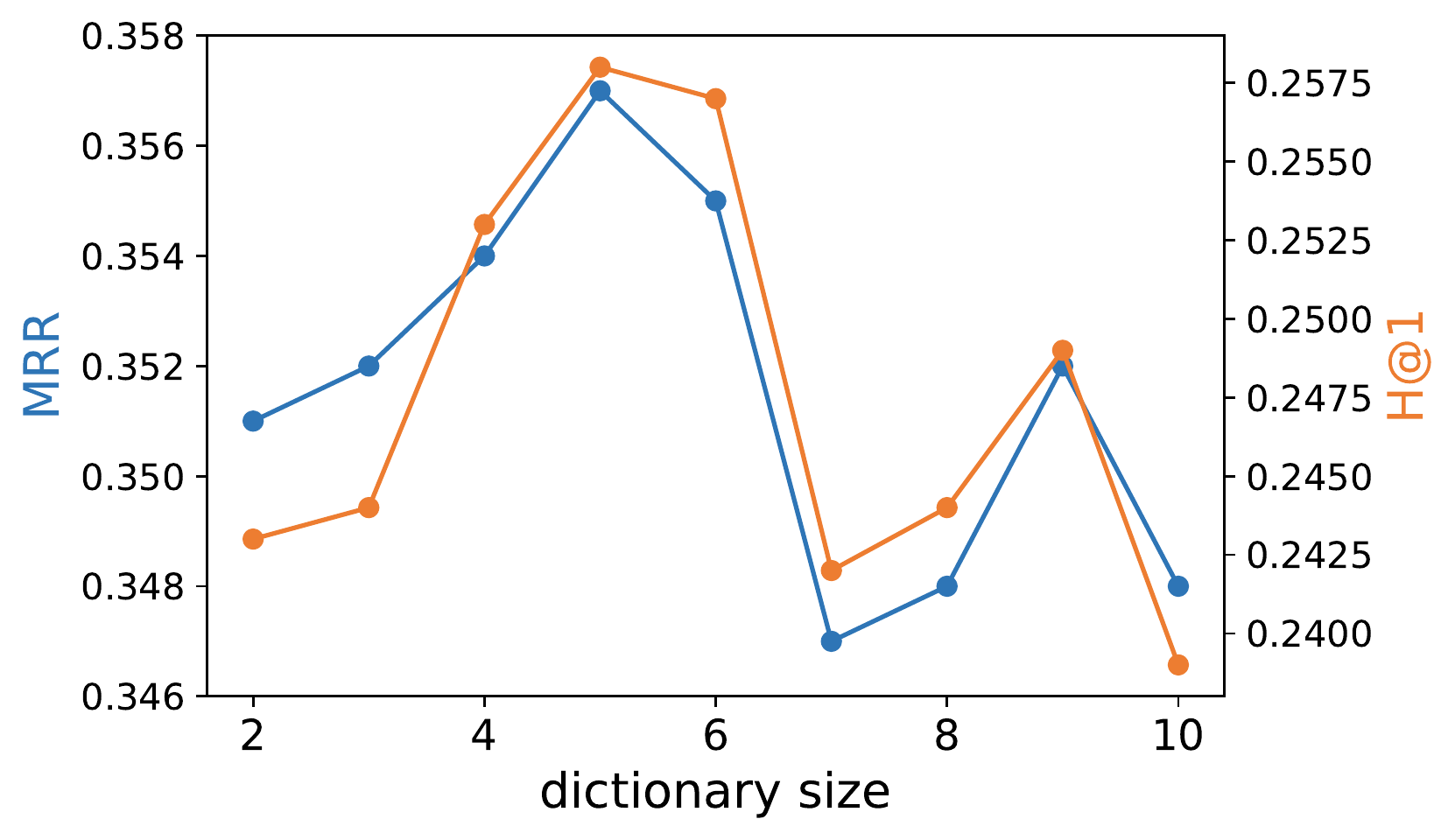}}
\subfloat[]{\includegraphics[width=0.33\hsize]{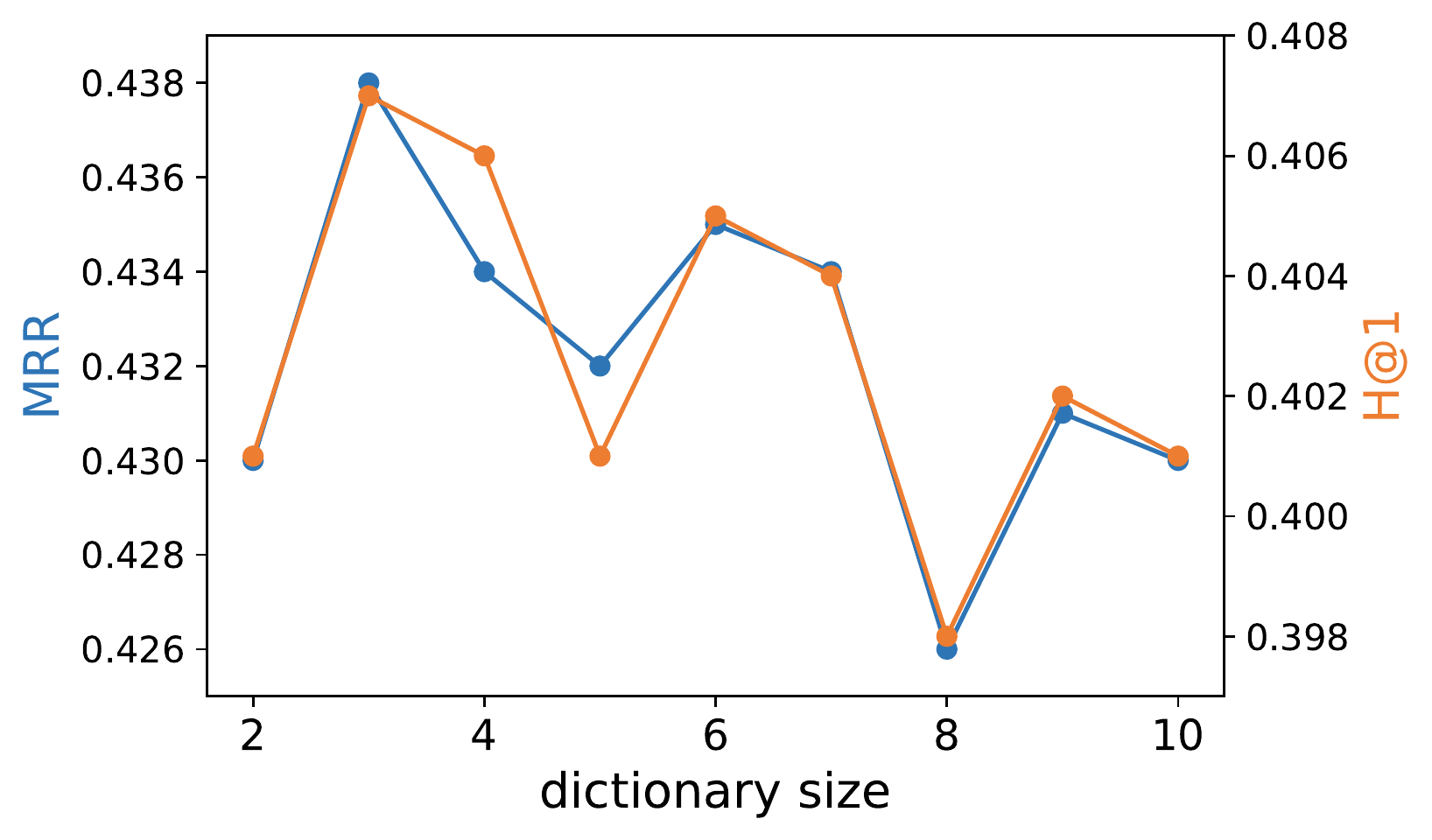}}
\subfloat[]{\includegraphics[width=0.33\hsize]{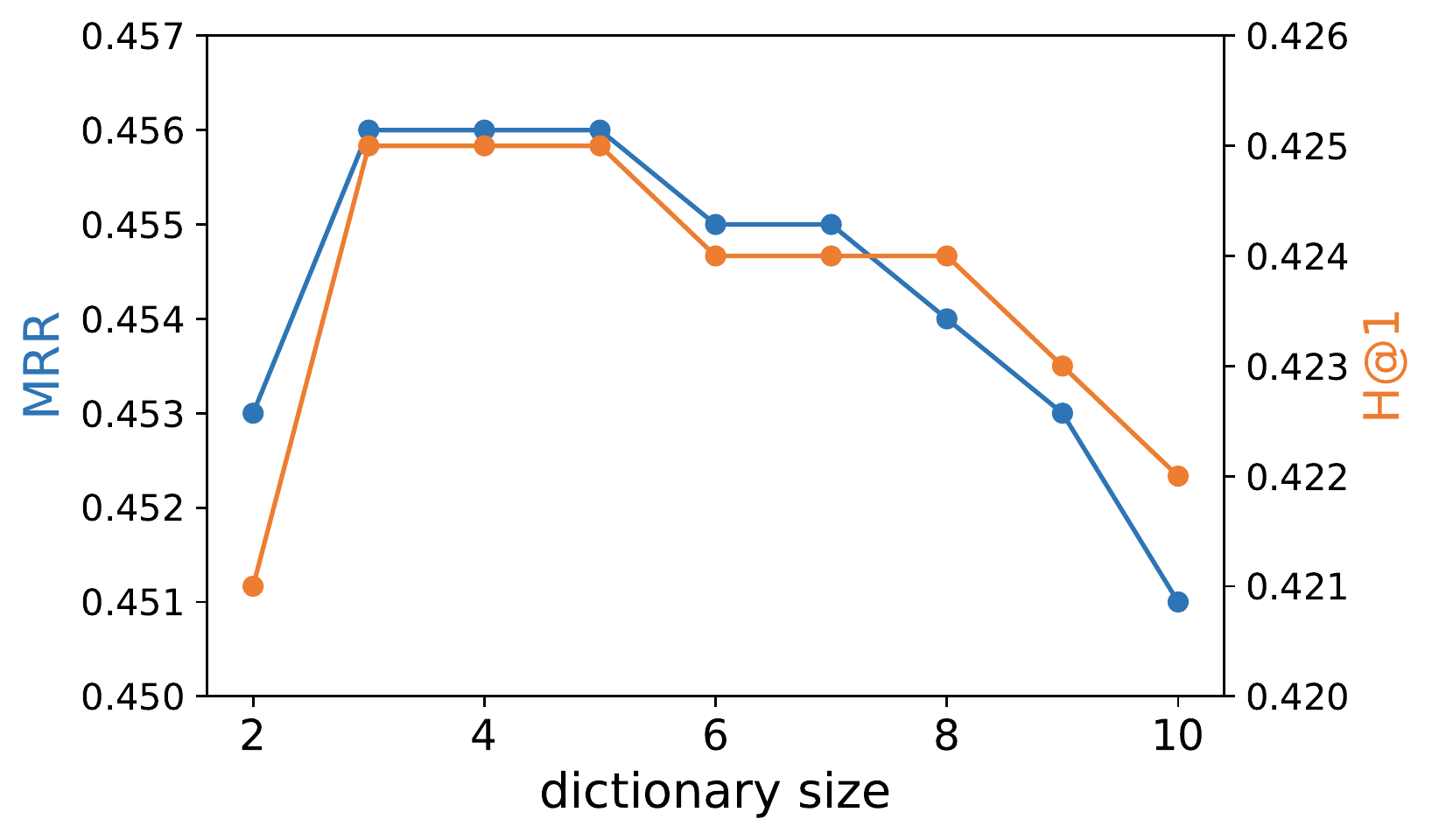}}
\caption{The changes of MRR and H@1 with the dictionary size $n$ increasing. (a) The CoDLR-TransE on FB15k237. (b) The CoDLR-DistMult on FB15k237. (c) The CoDLR-ConvE on FB15k237. (d) The CoDLR-TransE on WN18RR. (e) The CoDLR-DistMult on WN18RR. (f) The CoDLR-ConvE on WN18RR.}
\label{dict}
\end{figure*}

To better analyze how extended models benefit from CoDLR, we conduct following ablation studies. For the context module, we replace the context vector $\bm{C^h_r}$ with the central semantics embedding $\bm{\mathcal{D}^c_r}$ and the relation-connected entity embedding respectively to obtain \textbf{CSDLR} and \textbf{REDLR}. For the loss function with semantics consistency, we set $\lambda = 0$ to get \textbf{FSCoDLR}, which only optimizes the fine-grained semantics while training. 

The detailed link prediction results are reported in Table \ref{ablation}. It can be seen that after removing the context vector or semantic consistency constraint, the effects of models have declined to varying degrees, showing the importance of taking both entities and central semantics into consideration for the generation of lookup vector as well as keeping the semantic consistency constraint. Specifically, REDLR performs better than CSDLR in most cases, as the relation-connected entity contains more information than the central semantics under different questions, the information contained in the latter is actually fixed.

\subsection{Metrics Observation}

Although CoDLR shows superior performance on aforementioned experiments, the quality of dictionary lookup on which it relies has not been studied carefully. In this part, we re-examine the training process of CoDLR-TransE from the perspective of SOL, DIV and DAE, the results are shown in Fig.~\ref{metrics}, note that the average results over the whole datasets are plotted, and curves in Fig.~\ref{metrics}(c) are the absolute difference between the solid lines and the dashed lines of corresponding colors in Fig.~\ref{metrics}(b) essentially. 

For verifying the validity of lookup, we observe that the sparseness of lookup vectors rises first and then decreases slightly in Fig.~\ref{metrics}(a), which can be explained by the fact that one ambiguous relation may contain more than one fine-grained semantics, it's exactly the case that the previous clustering-based models fail to consider. In terms of the accuracy of dictionary, we find that despite the differences for the diversity of relation-connected entities and dictionaries reflected in Fig.~\ref{metrics}(b) under specific datasets, the DAE curves decrease with similar patterns in Fig.~\ref{metrics}(c), proving the stability of CoDLR in accuracy while dealing with different datasets.

\subsection{Parameter Study}

In this section, we study how the choices of the composition operator $\phi$ and the dictionary size $n$ affect the performance of CoDLR. Considering the great promotion of fine-grained semantics to H@1, we take MRR and H@1 as indicators to evaluate the model effects. For the composition operator $\phi$, the results in Fig.~\ref{contex} indicate that there does exist a best $\phi$ for different models and datasets. For the dictionary size $n$, the curves in Fig.~\ref{dict} all rise slowly when we increase $n$ from 2 to the optimal value, then the two indicators start falling when $n$ continues to increase, which can be attributed to the over-fitting caused by massive parameters.

\subsection{Case Study}

\begin{figure*}[!t]
	\centering
	\includegraphics[width=1.0\hsize]{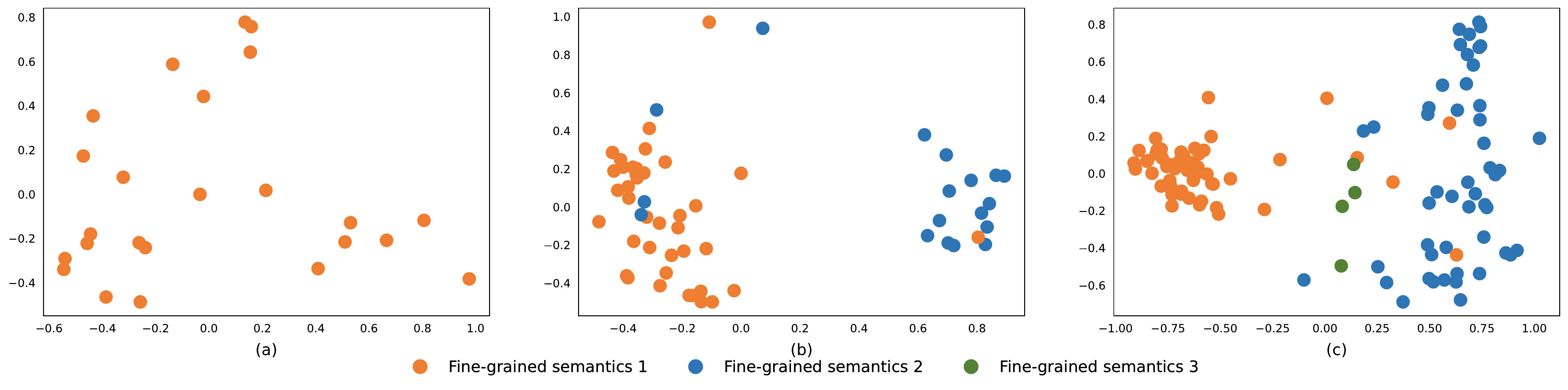}
	\caption{The number of fine-grained semantics of different relations under their corresponding entities is different. (a) The relation \textit{DesignFilm} has only one fine-grained semantics. (b) The relation \textit{SportsIn} has two fine-grained semantics (c) The relation \textit{LocationReligion} has three fine-grained semantics.}
	\label{casefig}
\end{figure*}

\begin{table*}[!t]
	\caption{Examples of fine-grained semantics for relations including \textit{DesignFilm}, \textit{SportsIn} and \textit{LocationReligion}.}
	\centering
	\begin{tabular}{l|l|l}
		\toprule
		Relation                           & Fine-grained semantics & Corresponding (head entity, tail entity) pairs  \\ \midrule
		\textit{DesignFilm}                & \textit{DesignFilm}                  & \textit{(Emile Kuri, The Heiress), (Marvin March, Flashdance)}  \\ \midrule
		\multirow{2}{*}{\textit{SportsIn}} & SummerSportsIn                       & \textit{(Synchronized swimming, France), (Racewalking, India)}  \\
		& WinterSportsIn                       & \textit{(Snowboarding, Austria), (Speed skating, Latvia)}          \\  \midrule
		\multirow{3}{*}{LocationReligion}  & CountryReligion                      & \textit{(Germany, Evangelicalism), (Angola, Protestantism)}        \\
		& StateReligion                        & \textit{(California, Churches of Christ), (Connecticut, Baptists)} \\
		& CityReligion                         & \textit{(Berlin, Protestantism), (Bali, Buddhism)}  \\ \bottomrule       
	\end{tabular}
	\label{casetab} 
\end{table*}

To demonstrate the ability of CoDLR to specifically handle fine-grained semantics with different granularity, we provide several convictive case studies in this part. Given one relation $r$, we reduce its corresponding entity embeddings $\bm{\mathcal{H}_r}$ to two-dimensional points by the principal component analysis (PCA) algorithm, then we set the index $i$ of the maximum value in the lookup vector for each point as its label like "fine-grained semantics $i$". The spatial distribution of above labeled points is shown in Fig.~\ref{casefig}. It can be seen that for relations with semantics in different granularity, the representation of the relation-connected entities with same label are located in the same area, indicating the strong adaptability and accuracy of contextual dictionary lookup. The detailed summary is shown in Table \ref{casetab}.

\section{Conclusion}
\label{cc}

In this work, we notice that the fine-grained semantics of relations play an important role while predicting missing links, and existing cluster-based models are limited by the two-stage training. To make up for the deficiency, CoDLR, an efficient end-to-end method, is proposed to enable the exploration of fine-grained semantics beyond conventional KGE models. In particular, CoDLR represents the relation with a dictionary and treats the composition of the center semantics and relation-connected entity as the context to generate a lookup vector, so that the fine-grained semantics can be determined from the dictionary, then both center and fine-grained semantics are optimized by the loss function to ensure their consistency. To thoroughly evaluate our proposed method, we extend different types of models including TransE, DistMult, and ConvE with CoDLR. The link prediction results on two well-established benchmarks demonstrate that they can surpass the original model as well as other competitors by a significant margin. Furthermore, we introduce two metrics, SOL and DAE, to evaluate the performance of the extended models. The results align with our expectations, confirming the effectiveness and accuracy of the dictionary lookup mechanism employed in CoDLR. 

Overall, our experimental findings highlight the superiority of CoDLR and its extended models in capturing fine-grained semantics and improving link prediction performance. However, due to time and resource constraints, we only extend several conventional models with simple structures in this work, and more advanced models are expected. For future work, we suggest two potential directions to further enhance the existing framework. On the one hand, the knowledge distillation can be combined with the existing framework, i.e., the scores predicted by central semantics can be regarded as the teacher's knowledge to guide the learning of fine-grained semantics. On the other hand, SOL and DAE, the two metrics can be added to the loss function as optimized terms to smooth the training.

\bibliographystyle{model6-num-names}
\bibliography{custom}

\begin{thebibliography}{10}
\expandafter\ifx\csname url\endcsname\relax
  \def\url#1{\texttt{#1}}\fi
\expandafter\ifx\csname urlprefix\endcsname\relax\def\urlprefix{URL }\fi
\expandafter\ifx\csname href\endcsname\relax
  \def\href#1#2{#2} \def\path#1{#1}\fi

\bibitem{qatask}
J.~Yin, X.~Jiang, Z.~Lu, L.~Shang, H.~Li, X.~Li,
  \href{http://www.ijcai.org/Abstract/16/422}{Neural generative question
  answering}, in: S.~Kambhampati (Ed.), Proceedings of the Twenty-Fifth
  International Joint Conference on Artificial Intelligence, {IJCAI} 2016, New
  York, NY, USA, 9-15 July 2016, {IJCAI/AAAI} Press, 2016, pp. 2972--2978.
\newline\urlprefix\url{http://www.ijcai.org/Abstract/16/422}

\bibitem{dstask}
Y.~Ma, P.~A. Crook, R.~Sarikaya, E.~Fosler{-}Lussier,
  \href{https://doi.org/10.1109/ICASSP.2015.7178992}{Knowledge graph inference
  for spoken dialog systems}, in: 2015 {IEEE} International Conference on
  Acoustics, Speech and Signal Processing, {ICASSP} 2015, South Brisbane,
  Queensland, Australia, April 19-24, 2015, {IEEE}, 2015, pp. 5346--5350.
\newblock \href {https://doi.org/10.1109/ICASSP.2015.7178992}
  {\path{doi:10.1109/ICASSP.2015.7178992}}.
\newline\urlprefix\url{https://doi.org/10.1109/ICASSP.2015.7178992}

\bibitem{irtask}
C.~Xiong, R.~Power, J.~Callan,
  \href{https://doi.org/10.1145/3038912.3052558}{Explicit semantic ranking for
  academic search via knowledge graph embedding}, in: R.~Barrett, R.~Cummings,
  E.~Agichtein, E.~Gabrilovich (Eds.), Proceedings of the 26th International
  Conference on World Wide Web, {WWW} 2017, Perth, Australia, April 3-7, 2017,
  {ACM}, 2017, pp. 1271--1279.
\newblock \href {https://doi.org/10.1145/3038912.3052558}
  {\path{doi:10.1145/3038912.3052558}}.
\newline\urlprefix\url{https://doi.org/10.1145/3038912.3052558}

\bibitem{rstask}
F.~Zhang, N.~J. Yuan, D.~Lian, X.~Xie, W.~Ma,
  \href{https://doi.org/10.1145/2939672.2939673}{Collaborative knowledge base
  embedding for recommender systems}, in: B.~Krishnapuram, M.~Shah, A.~J.
  Smola, C.~C. Aggarwal, D.~Shen, R.~Rastogi (Eds.), Proceedings of the 22nd
  {ACM} {SIGKDD} International Conference on Knowledge Discovery and Data
  Mining, San Francisco, CA, USA, August 13-17, 2016, {ACM}, 2016, pp.
  353--362.
\newblock \href {https://doi.org/10.1145/2939672.2939673}
  {\path{doi:10.1145/2939672.2939673}}.
\newline\urlprefix\url{https://doi.org/10.1145/2939672.2939673}

\bibitem{incompleteness}
X.~Dong, E.~Gabrilovich, G.~Heitz, W.~Horn, N.~Lao, K.~Murphy, T.~Strohmann,
  S.~Sun, W.~Zhang, \href{https://doi.org/10.1145/2623330.2623623}{Knowledge
  vault: a web-scale approach to probabilistic knowledge fusion}, in: S.~A.
  Macskassy, C.~Perlich, J.~Leskovec, W.~Wang, R.~Ghani (Eds.), The 20th {ACM}
  {SIGKDD} International Conference on Knowledge Discovery and Data Mining,
  {KDD} '14, New York, NY, {USA} - August 24 - 27, 2014, {ACM}, 2014, pp.
  601--610.
\newblock \href {https://doi.org/10.1145/2623330.2623623}
  {\path{doi:10.1145/2623330.2623623}}.
\newline\urlprefix\url{https://doi.org/10.1145/2623330.2623623}

\bibitem{transe}
A.~Bordes, N.~Usunier, A.~Garc{\'{\i}}a{-}Dur{\'{a}}n, J.~Weston, O.~Yakhnenko,
  \href{https://proceedings.neurips.cc/paper/2013/hash/1cecc7a77928ca8133fa24680a88d2f9-Abstract.html}{Translating
  embeddings for modeling multi-relational data} (2013) 2787--2795.
\newline\urlprefix\url{https://proceedings.neurips.cc/paper/2013/hash/1cecc7a77928ca8133fa24680a88d2f9-Abstract.html}

\bibitem{transh}
Z.~Wang, J.~Zhang, J.~Feng, Z.~Chen,
  \href{http://www.aaai.org/ocs/index.php/AAAI/AAAI14/paper/view/8531}{Knowledge
  graph embedding by translating on hyperplanes}, in: C.~E. Brodley, P.~Stone
  (Eds.), Proceedings of the Twenty-Eighth {AAAI} Conference on Artificial
  Intelligence, July 27 -31, 2014, Qu{\'{e}}bec City, Qu{\'{e}}bec, Canada,
  {AAAI} Press, 2014, pp. 1112--1119.
\newline\urlprefix\url{http://www.aaai.org/ocs/index.php/AAAI/AAAI14/paper/view/8531}

\bibitem{transr}
Y.~Lin, Z.~Liu, M.~Sun, Y.~Liu, X.~Zhu,
  \href{http://www.aaai.org/ocs/index.php/AAAI/AAAI15/paper/view/9571}{Learning
  entity and relation embeddings for knowledge graph completion}, in: B.~Bonet,
  S.~Koenig (Eds.), Proceedings of the Twenty-Ninth {AAAI} Conference on
  Artificial Intelligence, January 25-30, 2015, Austin, Texas, {USA}, {AAAI}
  Press, 2015, pp. 2181--2187.
\newline\urlprefix\url{http://www.aaai.org/ocs/index.php/AAAI/AAAI15/paper/view/9571}

\bibitem{transd}
G.~Ji, S.~He, L.~Xu, K.~Liu, J.~Zhao,
  \href{https://doi.org/10.3115/v1/p15-1067}{Knowledge graph embedding via
  dynamic mapping matrix}, in: Proceedings of the 53rd Annual Meeting of the
  Association for Computational Linguistics and the 7th International Joint
  Conference on Natural Language Processing of the Asian Federation of Natural
  Language Processing, {ACL} 2015, July 26-31, 2015, Beijing, China, Volume 1:
  Long Papers, The Association for Computer Linguistics, 2015, pp. 687--696.
\newblock \href {https://doi.org/10.3115/v1/p15-1067}
  {\path{doi:10.3115/v1/p15-1067}}.
\newline\urlprefix\url{https://doi.org/10.3115/v1/p15-1067}

\bibitem{rescal}
M.~Nickel, V.~Tresp, H.~Kriegel,
  \href{https://icml.cc/2011/papers/438\_icmlpaper.pdf}{A three-way model for
  collective learning on multi-relational data}, in: L.~Getoor, T.~Scheffer
  (Eds.), Proceedings of the 28th International Conference on Machine Learning,
  {ICML} 2011, Bellevue, Washington, USA, June 28 - July 2, 2011, Omnipress,
  2011, pp. 809--816.
\newline\urlprefix\url{https://icml.cc/2011/papers/438\_icmlpaper.pdf}

\bibitem{distmult}
B.~Yang, W.~Yih, X.~He, J.~Gao, L.~Deng,
  \href{http://arxiv.org/abs/1412.6575}{Embedding entities and relations for
  learning and inference in knowledge bases} (2015).
\newline\urlprefix\url{http://arxiv.org/abs/1412.6575}

\bibitem{complex}
T.~Trouillon, J.~Welbl, S.~Riedel, {\'{E}}.~Gaussier, G.~Bouchard,
  \href{http://proceedings.mlr.press/v48/trouillon16.html}{Complex embeddings
  for simple link prediction}, in: M.~Balcan, K.~Q. Weinberger (Eds.),
  Proceedings of the 33nd International Conference on Machine Learning, {ICML}
  2016, New York City, NY, USA, June 19-24, 2016, Vol.~48 of {JMLR} Workshop
  and Conference Proceedings, JMLR.org, 2016, pp. 2071--2080.
\newline\urlprefix\url{http://proceedings.mlr.press/v48/trouillon16.html}

\bibitem{quate}
S.~Zhang, Y.~Tay, L.~Yao, Q.~Liu,
  \href{https://proceedings.neurips.cc/paper/2019/hash/d961e9f236177d65d21100592edb0769-Abstract.html}{Quaternion
  knowledge graph embeddings} (2019) 2731--2741.
\newline\urlprefix\url{https://proceedings.neurips.cc/paper/2019/hash/d961e9f236177d65d21100592edb0769-Abstract.html}

\bibitem{nam}
Q.~Liu, H.~Jiang, Z.~Ling, S.~Wei, Y.~Hu,
  \href{http://arxiv.org/abs/1603.07704}{Probabilistic reasoning via deep
  learning: Neural association models}, CoRR abs/1603.07704 (2016).
\newblock \href {http://arxiv.org/abs/1603.07704} {\path{arXiv:1603.07704}}.
\newline\urlprefix\url{http://arxiv.org/abs/1603.07704}

\bibitem{conve}
T.~Dettmers, P.~Minervini, P.~Stenetorp, S.~Riedel,
  \href{https://www.aaai.org/ocs/index.php/AAAI/AAAI18/paper/view/17366}{Convolutional
  2d knowledge graph embeddings}, in: S.~A. McIlraith, K.~Q. Weinberger (Eds.),
  Proceedings of the Thirty-Second {AAAI} Conference on Artificial
  Intelligence, (AAAI-18), the 30th innovative Applications of Artificial
  Intelligence (IAAI-18), and the 8th {AAAI} Symposium on Educational Advances
  in Artificial Intelligence (EAAI-18), New Orleans, Louisiana, USA, February
  2-7, 2018, {AAAI} Press, 2018, pp. 1811--1818.
\newline\urlprefix\url{https://www.aaai.org/ocs/index.php/AAAI/AAAI18/paper/view/17366}

\bibitem{interacte}
S.~Vashishth, S.~Sanyal, V.~Nitin, N.~Agrawal, P.~P. Talukdar,
  \href{https://ojs.aaai.org/index.php/AAAI/article/view/5694}{Interacte:
  Improving convolution-based knowledge graph embeddings by increasing feature
  interactions}, in: The Thirty-Fourth {AAAI} Conference on Artificial
  Intelligence, {AAAI} 2020, The Thirty-Second Innovative Applications of
  Artificial Intelligence Conference, {IAAI} 2020, The Tenth {AAAI} Symposium
  on Educational Advances in Artificial Intelligence, {EAAI} 2020, New York,
  NY, USA, February 7-12, 2020, {AAAI} Press, 2020, pp. 3009--3016.
\newline\urlprefix\url{https://ojs.aaai.org/index.php/AAAI/article/view/5694}

\bibitem{rgcn}
M.~S. Schlichtkrull, T.~N. Kipf, P.~Bloem, R.~van~den Berg, I.~Titov,
  M.~Welling, \href{https://doi.org/10.1007/978-3-319-93417-4\_38}{Modeling
  relational data with graph convolutional networks}, in: A.~Gangemi,
  R.~Navigli, M.~Vidal, P.~Hitzler, R.~Troncy, L.~Hollink, A.~Tordai, M.~Alam
  (Eds.), The Semantic Web - 15th International Conference, {ESWC} 2018,
  Heraklion, Crete, Greece, June 3-7, 2018, Proceedings, Vol. 10843 of Lecture
  Notes in Computer Science, Springer, 2018, pp. 593--607.
\newblock \href {https://doi.org/10.1007/978-3-319-93417-4\_38}
  {\path{doi:10.1007/978-3-319-93417-4\_38}}.
\newline\urlprefix\url{https://doi.org/10.1007/978-3-319-93417-4\_38}

\bibitem{dog}
D.~Ruffinelli, S.~Broscheit, R.~Gemulla,
  \href{https://openreview.net/forum?id=BkxSmlBFvr}{You {CAN} teach an old dog
  new tricks! on training knowledge graph embeddings}, in: 8th International
  Conference on Learning Representations, {ICLR} 2020, Addis Ababa, Ethiopia,
  April 26-30, 2020, OpenReview.net, 2020.
\newline\urlprefix\url{https://openreview.net/forum?id=BkxSmlBFvr}

\bibitem{hrs}
Z.~Zhang, F.~Zhuang, M.~Qu, F.~Lin, Q.~He,
  \href{https://doi.org/10.18653/v1/d18-1358}{Knowledge graph embedding with
  hierarchical relation structure}, in: E.~Riloff, D.~Chiang, J.~Hockenmaier,
  J.~Tsujii (Eds.), Proceedings of the 2018 Conference on Empirical Methods in
  Natural Language Processing, Brussels, Belgium, October 31 - November 4,
  2018, Association for Computational Linguistics, 2018, pp. 3198--3207.
\newblock \href {https://doi.org/10.18653/v1/d18-1358}
  {\path{doi:10.18653/v1/d18-1358}}.
\newline\urlprefix\url{https://doi.org/10.18653/v1/d18-1358}

\bibitem{rotate}
Z.~Sun, Z.~Deng, J.~Nie, J.~Tang,
  \href{https://openreview.net/forum?id=HkgEQnRqYQ}{Rotate: Knowledge graph
  embedding by relational rotation in complex space} (2019).
\newline\urlprefix\url{https://openreview.net/forum?id=HkgEQnRqYQ}

\bibitem{ntn}
R.~Socher, D.~Chen, C.~D. Manning, A.~Y. Ng,
  \href{https://proceedings.neurips.cc/paper/2013/hash/b337e84de8752b27eda3a12363109e80-Abstract.html}{Reasoning
  with neural tensor networks for knowledge base completion}, in: C.~J.~C.
  Burges, L.~Bottou, Z.~Ghahramani, K.~Q. Weinberger (Eds.), Advances in Neural
  Information Processing Systems 26: 27th Annual Conference on Neural
  Information Processing Systems 2013. Proceedings of a meeting held December
  5-8, 2013, Lake Tahoe, Nevada, United States, 2013, pp. 926--934.
\newline\urlprefix\url{https://proceedings.neurips.cc/paper/2013/hash/b337e84de8752b27eda3a12363109e80-Abstract.html}

\bibitem{sme}
X.~Glorot, A.~Bordes, J.~Weston, Y.~Bengio,
  \href{http://arxiv.org/abs/1301.3485}{A semantic matching energy function for
  learning with multi-relational data} (2013).
\newline\urlprefix\url{http://arxiv.org/abs/1301.3485}

\bibitem{convkb}
D.~Q. Nguyen, T.~D. Nguyen, D.~Q. Nguyen, D.~Q. Phung,
  \href{https://doi.org/10.18653/v1/n18-2053}{A novel embedding model for
  knowledge base completion based on convolutional neural network}, in: M.~A.
  Walker, H.~Ji, A.~Stent (Eds.), Proceedings of the 2018 Conference of the
  North American Chapter of the Association for Computational Linguistics:
  Human Language Technologies, NAACL-HLT, New Orleans, Louisiana, USA, June
  1-6, 2018, Volume 2 (Short Papers), Association for Computational
  Linguistics, 2018, pp. 327--333.
\newblock \href {https://doi.org/10.18653/v1/n18-2053}
  {\path{doi:10.18653/v1/n18-2053}}.
\newline\urlprefix\url{https://doi.org/10.18653/v1/n18-2053}

\bibitem{convr}
X.~Jiang, Q.~Wang, B.~Wang,
  \href{https://doi.org/10.18653/v1/n19-1103}{Adaptive convolution for
  multi-relational learning}, in: J.~Burstein, C.~Doran, T.~Solorio (Eds.),
  Proceedings of the 2019 Conference of the North American Chapter of the
  Association for Computational Linguistics: Human Language Technologies,
  {NAACL-HLT} 2019, Minneapolis, MN, USA, June 2-7, 2019, Volume 1 (Long and
  Short Papers), Association for Computational Linguistics, 2019, pp. 978--987.
\newblock \href {https://doi.org/10.18653/v1/n19-1103}
  {\path{doi:10.18653/v1/n19-1103}}.
\newline\urlprefix\url{https://doi.org/10.18653/v1/n19-1103}

\bibitem{sacn}
C.~Shang, Y.~Tang, J.~Huang, J.~Bi, X.~He, B.~Zhou,
  \href{https://doi.org/10.1609/aaai.v33i01.33013060}{End-to-end
  structure-aware convolutional networks for knowledge base completion}, in:
  The Thirty-Third {AAAI} Conference on Artificial Intelligence, {AAAI} 2019,
  The Thirty-First Innovative Applications of Artificial Intelligence
  Conference, {IAAI} 2019, The Ninth {AAAI} Symposium on Educational Advances
  in Artificial Intelligence, {EAAI} 2019, Honolulu, Hawaii, USA, January 27 -
  February 1, 2019, {AAAI} Press, 2019, pp. 3060--3067.
\newblock \href {https://doi.org/10.1609/aaai.v33i01.33013060}
  {\path{doi:10.1609/aaai.v33i01.33013060}}.
\newline\urlprefix\url{https://doi.org/10.1609/aaai.v33i01.33013060}

\bibitem{vrgcn}
R.~Ye, X.~Li, Y.~Fang, H.~Zang, M.~Wang,
  \href{https://doi.org/10.24963/ijcai.2019/574}{A vectorized relational graph
  convolutional network for multi-relational network alignment}, in: S.~Kraus
  (Ed.), Proceedings of the Twenty-Eighth International Joint Conference on
  Artificial Intelligence, {IJCAI} 2019, Macao, China, August 10-16, 2019,
  ijcai.org, 2019, pp. 4135--4141.
\newblock \href {https://doi.org/10.24963/ijcai.2019/574}
  {\path{doi:10.24963/ijcai.2019/574}}.
\newline\urlprefix\url{https://doi.org/10.24963/ijcai.2019/574}

\bibitem{hole}
M.~Nickel, L.~Rosasco, T.~A. Poggio,
  \href{http://www.aaai.org/ocs/index.php/AAAI/AAAI16/paper/view/12484}{Holographic
  embeddings of knowledge graphs}, in: D.~Schuurmans, M.~P. Wellman (Eds.),
  Proceedings of the Thirtieth {AAAI} Conference on Artificial Intelligence,
  February 12-17, 2016, Phoenix, Arizona, {USA}, {AAAI} Press, 2016, pp.
  1955--1961.
\newline\urlprefix\url{http://www.aaai.org/ocs/index.php/AAAI/AAAI16/paper/view/12484}

\bibitem{freebase}
K.~D. Bollacker, C.~Evans, P.~K. Paritosh, T.~Sturge, J.~Taylor,
  \href{https://doi.org/10.1145/1376616.1376746}{Freebase: a collaboratively
  created graph database for structuring human knowledge}, in: J.~T. Wang
  (Ed.), Proceedings of the {ACM} {SIGMOD} International Conference on
  Management of Data, {SIGMOD} 2008, Vancouver, BC, Canada, June 10-12, 2008,
  {ACM}, 2008, pp. 1247--1250.
\newblock \href {https://doi.org/10.1145/1376616.1376746}
  {\path{doi:10.1145/1376616.1376746}}.
\newline\urlprefix\url{https://doi.org/10.1145/1376616.1376746}

\bibitem{wordnet}
G.~A. Miller, \href{http://doi.acm.org/10.1145/219717.219748}{Wordnet: {A}
  lexical database for english}, Commun. {ACM} 38~(11) (1995) 39--41.
\newblock \href {https://doi.org/10.1145/219717.219748}
  {\path{doi:10.1145/219717.219748}}.
\newline\urlprefix\url{http://doi.acm.org/10.1145/219717.219748}

\bibitem{nell}
A.~Carlson, J.~Betteridge, B.~Kisiel, B.~Settles, E.~R.~H. Jr., T.~M. Mitchell,
  \href{http://www.aaai.org/ocs/index.php/AAAI/AAAI10/paper/view/1879}{Toward
  an architecture for never-ending language learning}, in: M.~Fox, D.~Poole
  (Eds.), Proceedings of the Twenty-Fourth {AAAI} Conference on Artificial
  Intelligence, {AAAI} 2010, Atlanta, Georgia, USA, July 11-15, 2010, {AAAI}
  Press, 2010.
\newline\urlprefix\url{http://www.aaai.org/ocs/index.php/AAAI/AAAI10/paper/view/1879}

\bibitem{yago3}
F.~Mahdisoltani, J.~Biega, F.~M. Suchanek,
  \href{http://cidrdb.org/cidr2015/Papers/CIDR15\_Paper1.pdf}{{YAGO3:} {A}
  knowledge base from multilingual wikipedias}, in: Seventh Biennial Conference
  on Innovative Data Systems Research, {CIDR} 2015, Asilomar, CA, USA, January
  4-7, 2015, Online Proceedings, www.cidrdb.org, 2015.
\newline\urlprefix\url{http://cidrdb.org/cidr2015/Papers/CIDR15\_Paper1.pdf}

\bibitem{fb15k237}
K.~Toutanova, D.~Chen, Observed versus latent features for knowledge base and
  text inference, in: Proceedings of the 3rd workshop on continuous vector
  space models and their compositionality, 2015, pp. 57--66.

\bibitem{mrr}
W.~Zhang, B.~Paudel, W.~Zhang, A.~Bernstein, H.~Chen,
  \href{https://doi.org/10.1145/3289600.3291014}{Interaction embeddings for
  prediction and explanation in knowledge graphs}, in: J.~S. Culpepper,
  A.~Moffat, P.~N. Bennett, K.~Lerman (Eds.), Proceedings of the Twelfth {ACM}
  International Conference on Web Search and Data Mining, {WSDM} 2019,
  Melbourne, VIC, Australia, February 11-15, 2019, {ACM}, 2019, pp. 96--104.
\newblock \href {https://doi.org/10.1145/3289600.3291014}
  {\path{doi:10.1145/3289600.3291014}}.
\newline\urlprefix\url{https://doi.org/10.1145/3289600.3291014}

\bibitem{reevaluation}
Z.~Sun, S.~Vashishth, S.~Sanyal, P.~P. Talukdar, Y.~Yang,
  \href{https://doi.org/10.18653/v1/2020.acl-main.489}{A re-evaluation of
  knowledge graph completion methods}, in: D.~Jurafsky, J.~Chai, N.~Schluter,
  J.~R. Tetreault (Eds.), Proceedings of the 58th Annual Meeting of the
  Association for Computational Linguistics, {ACL} 2020, Online, July 5-10,
  2020, Association for Computational Linguistics, 2020, pp. 5516--5522.
\newblock \href {https://doi.org/10.18653/v1/2020.acl-main.489}
  {\path{doi:10.18653/v1/2020.acl-main.489}}.
\newline\urlprefix\url{https://doi.org/10.18653/v1/2020.acl-main.489}

\bibitem{pytorch}
A.~Paszke, S.~Gross, F.~Massa, A.~Lerer, J.~Bradbury, G.~Chanan, T.~Killeen,
  Z.~Lin, N.~Gimelshein, L.~Antiga, A.~Desmaison, A.~K{\"{o}}pf, E.~Z. Yang,
  Z.~DeVito, M.~Raison, A.~Tejani, S.~Chilamkurthy, B.~Steiner, L.~Fang,
  J.~Bai, S.~Chintala,
  \href{https://proceedings.neurips.cc/paper/2019/hash/bdbca288fee7f92f2bfa9f7012727740-Abstract.html}{Pytorch:
  An imperative style, high-performance deep learning library} (2019)
  8024--8035.
\newline\urlprefix\url{https://proceedings.neurips.cc/paper/2019/hash/bdbca288fee7f92f2bfa9f7012727740-Abstract.html}

\bibitem{adam}
D.~P. Kingma, J.~Ba, \href{http://arxiv.org/abs/1412.6980}{Adam: {A} method for
  stochastic optimization} (2015).
\newline\urlprefix\url{http://arxiv.org/abs/1412.6980}

\end{thebibliography}

\end{document}